\documentclass{bmvc2k}

\usepackage[capitalize,noabbrev]{cleveref}
\usepackage{booktabs} 
\usepackage{soul}

\newtoggle{editmode}
\togglefalse{editmode} 

\newcommand{\add}[1]{\iftoggle{editmode}{\textcolor{blue}{#1}}{#1}}

\usepackage{caption}
\newcommand{\eg}[1]{e.g.,}

\newcommand{\ddim}[1]{DDIM #1}
\newcommand{\xa}[1]{$x^{(A)}_{#1}$}
\newcommand{\xb}[1]{$x^{(B)}_{#1}$}
\newcommand{\za}[1]{$z^{(A)}_{#1}$}
\newcommand{\zb}[1]{$z^{(B)}_{#1}$}

\newcommand{\pred}[2]{\hat{#1}_{#2}}

\title{Seed-to-Seed: Unpaired Image Translation in Diffusion Seed Space}

\addauthor{Or Greenberg}{or.greenberg@gm.com}{1,2}
\addauthor{Eran Kishon}{eran.kishon@gm.com}{1}
\addauthor{Dani Lischinski}{danix3d@gmail.com}{2}

\addinstitution{
 General Motors R\&D\\
}
\addinstitution{
 The Hebrew University of Jerusalem\\
 Jerusalem, Israel\\
}

\runninghead{Greenberg, Kishon, Lischinski}{Seed-to-Seed Translation}

\def\eg{\emph{e.g}\bmvaOneDot}

\begin{document}

\maketitle

\begin{abstract}
We introduce Seed-to-Seed Translation (StS), a novel approach that combines GANs and diffusion models (DMs) for unpaired Image-to-Image Translation. Our approach is aimed at global translations of complex automotive scenes, where close adherence to the structure and semantics of the source image is essential. 
We demonstrate that the semantic information encoded in the space of inverted latents (seeds) of a pretrained DM, dubbed as the \emph{seed-space}, can be used for discriminative tasks, and leverage this information to perform image-to-image translation. Our method involves training an \emph{sts-GAN}, an unpaired seed-to-seed translation model, based on CycleGAN. The translated seeds are used as the starting point for the DM's sampling process, while structure preservation is ensured using a ControlNet. We demonstrate the effectiveness of our approach for structure-preserving translation of complex automotive scenes, showcasing superior performance compared to existing GAN-based and diffusion-based methods. 
In addition to advancing the SoTA in automotive scene translations, our approach offers a fresh perspective on leveraging the semantic information encoded within the seed-space of pretrained DMs for effective image editing and manipulation.
\end{abstract}

\section{Introduction}
\label{sec:intro}

\begin{figure}
\centering
\includegraphics[width=0.89\textwidth]{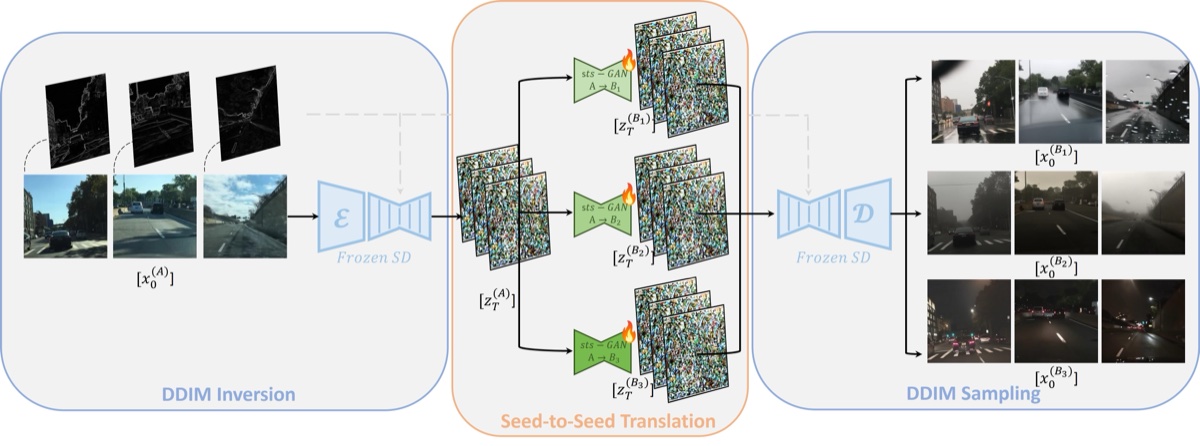}
  \caption{\textbf{Seed-to-Seed Translation} addresses the unpaired Image-to-Image Translation task from the source domain $A$ to some target domain $B_i$ by performing the translation in the seed-space of a pretrained diffusion model. The source image \xa{0} is first inverted to a corresponding seed \za{T}. Then the initial seed is translated to a target domain referred seed \zb{T}, which is finally sampled to yield the target domain output \xb{0}. Here we demonstrate a translation from the source domain ``clear day" to 3 different target domains: ``rainy day", ``foggy day" and ``clear night" denoted $B_1, B_2$ and $B_3$, respectively.
  }
\label{fig:flowchart}
\end{figure}

Diffusion Models (DMs) have emerged as powerful generative tools, synthesizing images by iteratively transforming noise samples, commonly referred as \emph{``seeds''}, into images \cite{ho2020denoising,sohl2015deep,song2020denoising}. State-of-the-art text-to-image diffusion models, \eg,~\cite{ramesh2022hierarchical,saharia2022photorealistic,rombach2022high} generate diverse and photo-realistic content, prompting efforts to repurpose DMs for a wide range of image editing tasks.

One such task is Image-to-Image Translation (I2IT), where an image is converted from one domain to another while preserving various aspects. In fact, I2IT encompasses a variety of tasks that differ in their required adherence to the source image. For instance, day-to-night translation demands perfect adherence to the structure of the original image, altering only appearance. In contrast, cat-to-dog translations might only preserve the pose and/or fur colors, while allowing other details to change.

Despite their generative power, DM-based editing methods often struggle with I2IT tasks requiring strict preservation of complex structures, such as those found in automotive scenes. Consequently, most current approaches in this domain rely on GANs \cite{goodfellow2014generative}, which offer stronger structural guarantees but lack the generative richness of DMs. 

In DM-based editing, manipulations often begin from inverted latents—e.g., via DDIM inversion \cite{song2020denoising}. \add{In this work, we follow the common convention and denote the initial latent as a \emph{seed}, even when it is obtained by inverting an image rather than by random sampling.} These seeds encode both attributes to be retained and those to be changed, making it difficult to isolate domain-specific elements. Moreover, editing typically occurs along the DM sampling trajectory, spanning multiple latent spaces. This further complicates the separation of domain-specific and agnostic features at each step.

To address these challenges, we introduce a novel approach to unpaired I2IT that operates directly in the \emph{seed-space} of a pretrained DM, prior to sampling. We show that inverted seeds encode rich semantic information and allow meaningful manipulations within this space. Our method enables controlled translations that preserve structural content while modifying domain-specific features.

We refer to this process as \emph{Seed-to-Seed Translation} (StS), implemented using a GAN-based translator reffered as \emph{sts-GAN}, trained on pairs of inverted seeds from the source and target domains. At inference, an input image is inverted to a seed, translated using \emph{sts-GAN}, and sampled via the DM to produce the target image (\Cref{fig:flowchart}). A ControlNet \cite{zhang2023adding} ensures the structural integrity of the source image is preserved during sampling.

Our focus is on translations that require high structural fidelity—particularly in complex automotive scenes—while enabling substantial changes in appearance. For example, in day-to-night translation, the geometry and layout must be preserved while lighting and atmospheric effects change. In photo restoration, structure and identity must remain intact even if appearance varies significantly. We demonstrate our method on automotive scene translations, including day-to-night and weather changes. Unlike GANs, which often fail to produce realistic target domain content, or DMs, which may distort structure, our approach maintains geometric integrity while achieving high realism. In the supplementary material, we also explore the applicability of our approach for several 
nonautomotive unpaired image translation tasks.

\noindent
We summarize our main contributions as follows:
\begin{enumerate}
    \item A fresh perspective on the semantic content of the seeds inverted by a pre-trained DMs, and the structure of the seed-space defined by them, highlighting the potential of image manipulations within the seed space, rather than along the sampling trajectory.
    \item We propose StS: a novel method for unpaired, structure-preserving domain translation in complex scenes.
    \item A hybrid framework combining GAN-based translation with diffusion-based generation, leveraging their complementary strengths.
\end{enumerate}    
\section{Related Work}
\label{sec:RelatedWork}

\textbf{Unpaired I2IT.} Unpaired Image-to-Image Translation (I2IT) focuses on translating images across domains without paired training examples. It has gained traction for applications like style transfer \cite{zhu2017unpaired, liu2017unsupervised, huang2018multimodal, jiang2020tsit, dutta2022seeing}, semantic segmentation \cite{guo2020gan, wu2021semi, kang2021domain}, and image enhancement \cite{chen2018deep, altakrouri2021image}. Many unpaired I2IT methods are GAN-based \cite{goodfellow2014generative}, using cycle consistency \cite{zhu2017unpaired, hoffman2017cycada} to preserve structure during translation. This regularization mitigates mode collapse and encourages content preservation.

\textbf{Diffusion-based Editing.} Diffusion models (DMs) \cite{ho2020denoising, dhariwal2021diffusion, rombach2022high} have enabled various image editing tasks. Local editing approaches, such as using masks or attention maps \cite{couairon2022diffedit, parmar2023zero, hertz2022prompt}, yield strong results but lack the global consistency needed for full-scene translations. In contrast, methods like Imagic \cite{kawar2023imagic} allow non-rigid edits but often sacrifice structural fidelity.

Globally-constrained methods like SDEdit \cite{meng2021sdedit} and PnP \cite{tumanyan2023plug} inject source image information early in the generation process, achieving better structure preservation but facing trade-offs between realism and faithfulness. Layout-to-image approaches \cite{zhang2023adding, mou2023t2iadapter, li2025controlnet} offer diverse outputs by conditioning on low-dimensional spatial cues (e.g., depth, edges), though they lack the detail required for precise I2IT.

\textbf{Seed Manipulations.} Most methods initiate sampling from a fixed or inverted seed and modify images during denoising. SeedSelect \cite{samuel2023all} instead optimizes the seed itself to produce rare objects via backpropagation through the entire sampling process. In contrast, our method performs seed-level optimization for unpaired I2IT, operating directly in seed-space rather than along the sampling trajectory.

\section{Method}
\label{sec:method}

In this section we introduce \emph{StS}, an image translation model that operates directly in the seed-space of a pretrained diffusion model. We begin by exploring the meaningfulness of the seed-space and the ability to access the information encoded within the seeds (\Cref{subsec:meaning}). Next, in \Cref{subsec:PaS} we show how seed meaningfulness may be leveraged to perform unpaired image translation within the seed-space using our proposed \emph{StS} model.

\subsection{Meaningful Seed Space}
\label{subsec:meaning}

Diffusion models \cite{ho2020denoising,sohl2015deep,song2019generative} generate images by mapping Gaussian noise (seeds) to images via a stochastic process. To edit real images, one must invert them into the model’s seed-space \cite{wallace2023edict, parmar2023zero, mokady2023null, tumanyan2023plug}. We adopt DDIM's deterministic sampling and inversion processes \cite{song2020denoising}; formal details are provided in the supplementary material. Deterministic \ddim{sampling} defines an injective mapping from seed-space to image-space. Similarly, \ddim{inversion} maps images back to seed-space. 

The DM's backbone iteratively decodes the seed across the diffusion steps \cite{ho2020denoising}. Editing methods typically intervene during this decoding by fine-tuning the decoder's weights \cite{kawar2023imagic, ruiz2023dreambooth}, modifying the decoder's condition input \cite{meng2021sdedit, zhang2023adding}, or injecting cross-attention elements across processes of different images \cite{tumanyan2023plug, hertz2022prompt, parmar2023zero}. In all cases, edits occur during the transformation of the latent seed into image.

\begin{figure}
  \centering
\includegraphics[width=0.8\linewidth]{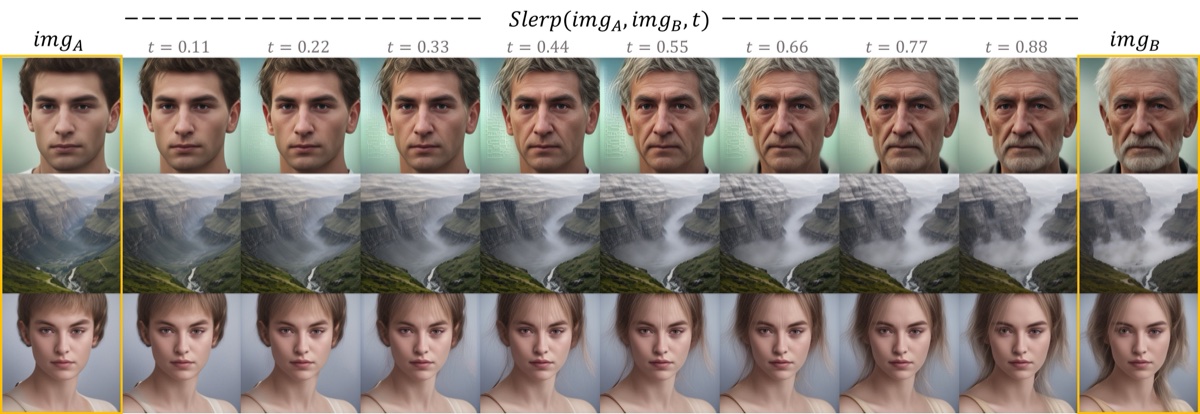}
  \caption{\textbf{Inverted seed interpolation.} Spherical interpolation between DDIM-inverted seeds from $\textit{img}_A$ and $\textit{img}_B$ yields a semantically coherent transformation between the images.}
  \label{fig:seed_inter}
\end{figure}

Prior work \cite{song2020denoising} shows seed interpolation yields smooth image transitions. As illustrated in \Cref{fig:seed_inter}, spherical interpolation (slerp \cite{shoemake1985animating}) between two inverted seeds produces semantically meaningful transitions: e.g., aging (row 1), fog density (row 2), and hair length (row 3). This suggests that seed-space is structurally informative and supports meaningful semantic operations. For instance, the young man (first row) appears progressively older as the interpolation parameter $t$ approaches $1$. Similarly, this gradual transformation is reflected in the increasing fog density and changing hair length in the second and third rows, respectively. This illustrates that seed-space encodes structured, interpretable information. To quantify this, we train ResNet18 \cite{he2016deep} classifiers on both images and their DDIM-inverted seeds (using Stable Diffusion 2.1 \cite{rombach2022high}) across classification tasks. As shown in \Cref{tab:class}, seed-based classifiers perform nearly as well as image-based ones, confirming that seed-space retains significant semantic information across scene (time of day), object (dog/cat), and sub-object (age) level attributes. In this work, we embrace this observation and further leverage this structure to perform image translation within the \ddim{inverted} seed space directly, before sampling.

\begin{table}
\centering

  \begin{tabular}{l|c|c}
      Task &  seeds &  images \\
       \hline \hline
       Day/Night    &  98.37\% &  98.47\% \\
       Cat/Dog      &  90.10\% &  98.53\% \\
       Older/Younger    &  92.60\% &  97.90\% \\
      \hline
  \end{tabular}
  \caption{\textbf{Classifier Accuracy Comparison.} Classifiers are trained once on image inputs and once on their corresponding inverted seeds. The tasks are day/night, cat/dog, and older/younger (using the \emph{BDD100k} \cite{yu2020bdd100k}, \emph{AFHQ} \cite{choi2020stargan}, and \emph{FFHQ} \cite{karras2019style} datasets, respectively). More details can be found in the supplementary material.}
  \label{tab:class}
\end{table}

\subsection{\emph{StS}: I2IT in Diffusion Seed Space}
\label{subsec:PaS}

We aim to perform unpaired I2IT within the seed-space of a pre-trained diffusion model by leveraging the information encoded in the \ddim{inverted} seeds. Consequently, we train a dedicated translation model that learns a mapping between seeds corresponding to images from a source domain \emph{A} to seeds corresponding to images from a target domain \emph{B}. We train our network, referred to as \emph{sts-GAN}, over a set of DDIM-inverted seeds from the source and target domains, using the CycleGAN architecture \cite{zhu2017unpaired} and training strategy. \add{clasiffier-free guidance (CFG)} scale $\omega = 1.0$ is used to accurately invert the unpaired source and target domain training images to the seed-space.

\Cref{fig:flowchart} presents a diagram depicting our method. At inference time, we first encode the input source image \xa{0} to the Stable Diffusion (SD) latent space, yielding \za{0}, and apply \ddim{inversion} (with a source-domain-referred prompt) to obtain a seed \za{T}. Next, we translate \za{T} to a target-domain-referred seed \zb{T} using our \emph{sts-GAN}. Finally, we sample \zb{T} using the same pre-trained SD model (with a target-domain-referred prompt), yielding the final denoised code \zb{0}, which is decoded to the resulting image \xb{0}.

\begin{figure}
  \centering
\includegraphics[width=0.8\columnwidth]{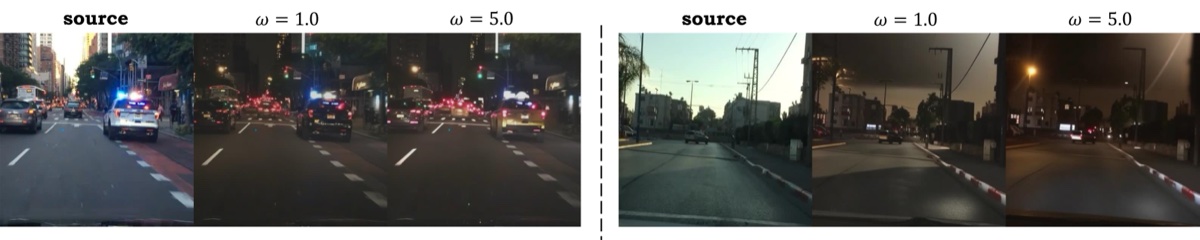}
  \caption{\textbf{Day-to-night translation with \emph{StS} using different CFG-scales.} While achieving a global night-time appearance, a low CFG-scale ($\omega = 1$) may result in lack of local domain-related semantic effects (middle). Using a higher CFG-scale ($\omega = 5$) introduces these important effects (right). 
  The same prompt ``A clear night'' is used in both columns.
  }
  \label{fig:ablation_cond}
\end{figure}

While \emph{sts-GAN} successfully translates source-referred seeds into target-referred ones, \ddim{sampling} these seeds without CFG typically results in images suffering from a lack of local semantic effects, despite the use of a target domain-referred prompt (\eg, ``A clear night'', for the day-to-night translation). For example, as demonstrated in the ``$\omega=1.0$'' columns of \Cref{fig:ablation_cond}, a day-to-night translation of automotive images might lack car lights, street lights, and reflections (left), or retain some daytime-like shadows on the road surface (right). To encourage such domain-specific effects, we employ CFG with $\omega = 5.0$, in conjunction with the same target-referred prompt. 

The cyclic consistency mechanism employed during \emph{sts-GAN} training enforces structural similarity between the source and the output \emph{within the seed space}. However, this similarity might not be maintained as the translated seed \zb{T} is sampled back to the image space. This issue becomes more pronounced when using CFG, as the extrapolation amplifies the accumulated errors from the \ddim{inversion} \cite{mokady2023null}. Consequently, even if the translation from \za{T} to \zb{T} is perfect in seed space, the final image \xb{0} may significantly deviate from the structure and content of the source image. To address this, we employ ControlNet \cite{zhang2023adding} to enforce structural similarity between the source image and the final output throughout the sampling trajectory.

The ``$\omega=5.0$'' column of \Cref{fig:ablation_cond} demonstrates that spatially-guided conditional sampling enhances the target-domain appearance, introducing the missing effects, while remaining faithful to the source image's structure. \add{Additional discussion and quantitative evaluation of the CFG scale factor are provided in the supplementary material.}

\section{Experiments}
\label{sec:Experiments}

We evaluate our method through extensive experiments on automotive unpaired image translation tasks, comparing against leading GAN-based and globally-constrained DM-based approaches. While GAN-based methods are currently considered state-of-the-art for lighting and weather translation in automotive scenes, our results demonstrate the efficacy of diffusion models in these challenging tasks. We present quantitative results on the Day-to-Night task and qualitative results across multiple image translation tasks, followed by an ablation study of our method's components. Code and models will be available upon publication.

\subsection{Implementation Details}

We evaluate unpaired I2IT tasks using the Berkeley DeepDrive \emph{BDD100k} \cite{yu2020bdd100k} and \emph{DENSE} \cite{bijelic2020seeing} datasets. Our implementation uses Stable Diffusion (SD) 2.1 \cite{rombach2022high} at $512\times512$ resolution as the diffusion backbone. For \emph{sts-GAN}, we employ a modified \emph{ResNet18} encoder with 4-channel input to match SD's latent space, omitting the final normalization layer. We trained \emph{sts-GAN} following the methodology of \citet{zhu2017unpaired}. Note that due to the low dimensionality of SD's latent space (8 times smaller than the image space along each axis, with an additional channel), training \emph{sts-GAN} in the latent space is significantly faster than training GANs in the image space. Due to SD's limited performance on automotive datasets, we finetune both SD 2.1 and its corresponding ControlNet \cite{zhang2023adding} on the \emph{BDD100k} training set using Diffusers' \cite{von-platen-etal-2022-diffusers} default scheme. Both \ddim{sampling} and \ddim{inversion} use 20 timesteps, with CFG-scales of $\omega = 1.0$ for inversion and $\omega = 5.0$ for forward sampling.

\subsection{Baselines}

We compare our Day-to-Night translation on \emph{BDD100k} against GAN-based methods (CycleGAN \cite{zhu2017unpaired}, MUNIT \cite{huang2018multimodal}, TSIT \cite{jiang2020tsit}, AU-GAN \cite{kwak2021adverse}, and CycleGAN-Turbo \cite{parmar2024one}), using the provided day2night checkpoints for AU-GAN and CycleGAN-Turbo. We trained the other models for 100 epochs using the provided public code with default hyperparameters, and selected the best checkpoints.

As mentioned in \Cref{sec:RelatedWork}, while state-of-the-art image editing techniques deliver outstanding results for object-level edits or relatively straightforward images, they often fail when applied to global edits of complex scenes. Diffusion-based methods, in particular, struggle to balance high fidelity to the source image with achieving the desired modifications in such challenging scenarios. Qualitative examples of these limitations are provided in the supplementary material. For our comparisons, we selected diffusion-based baselines with global-constraints, which are more suitable for global edits (as discussed in \Cref{sec:RelatedWork}). Specifically, we quantitatively evaluate our performance against SDEdit \cite{meng2021sdedit} with varying strength parameters ($0.5$, $0.7$, $0.9$) and Plug-and-Play (PnP) \cite{tumanyan2023plug}. We also include comparisons to a pure ControlNet \cite{zhang2023adding}, which is designed for image synthesis using a combination of textual and spatial conditions. To ensure a fair comparison, we utilize our fine-tuned U-Net for zero-shot diffusion-based methods (SDEdit and PnP) as well as ControlNet when working with the automotive datasets. For all these methods, we use the default settings of 50 timesteps and CFG-scale $\omega = 7.5$ during inference.

\subsection{Evaluation Metrics}

We follow the standard evaluation protocol used in prior GAN-based I2IT works \cite{brock2018large, liang2021high, liu2019learning}, employing SSIM  \cite{wang2004image} and FID \cite{frechet1957distance, heusel2017gans} to assess weather and lighting translation tasks. While feature-based metrics like DINO-Struct-Dist \cite{tumanyan2022splicing} have gained popularity, we found them unstable for complex automotive scenes. Given the limited size of our validation datasets (up to a few thousands samples per domain), we additionally report KID \cite{binkowski2018demystifying} and MMD \cite{gretton2012kernel}, which are considered more suitable for smaller datasets.

\subsection{Results}

\begin{figure}
    \centering
    \begin{subfigure}
        \centering
        \includegraphics[width=0.9\linewidth]{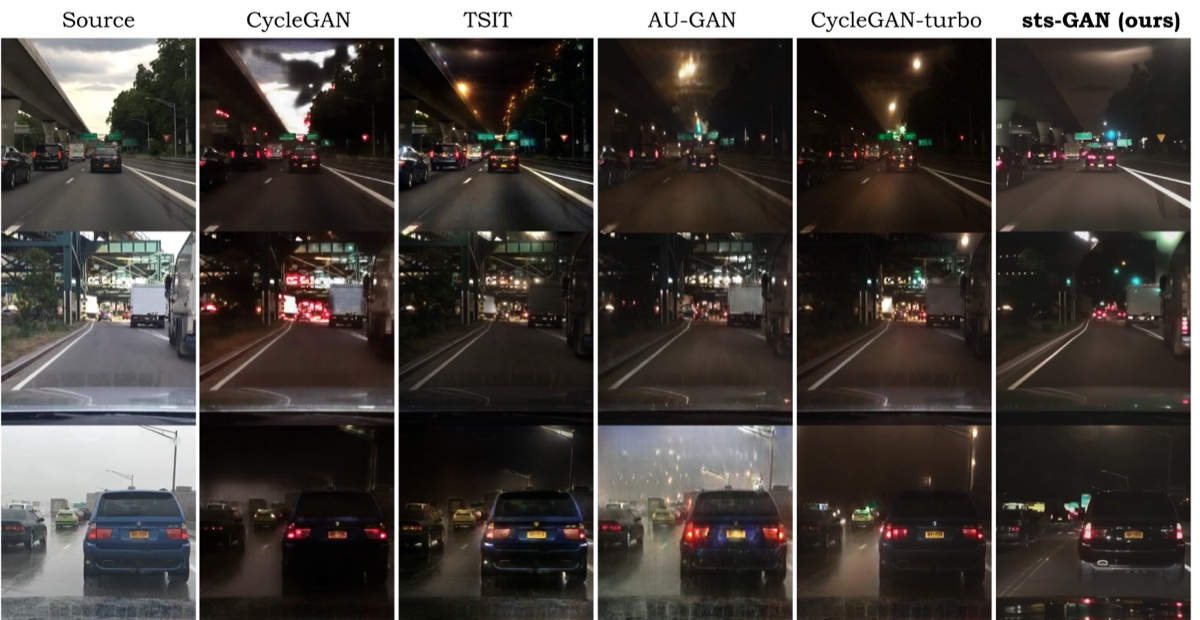}
        \caption{Qualitative comparison for Day-to-Night translation over the \emph{BDD100k} dataset - GAN-based baselines}
        \label{subfig:GAN}
    \end{subfigure}
    \vfill
    \begin{subfigure}
        \centering
        \includegraphics[width=0.9\linewidth]{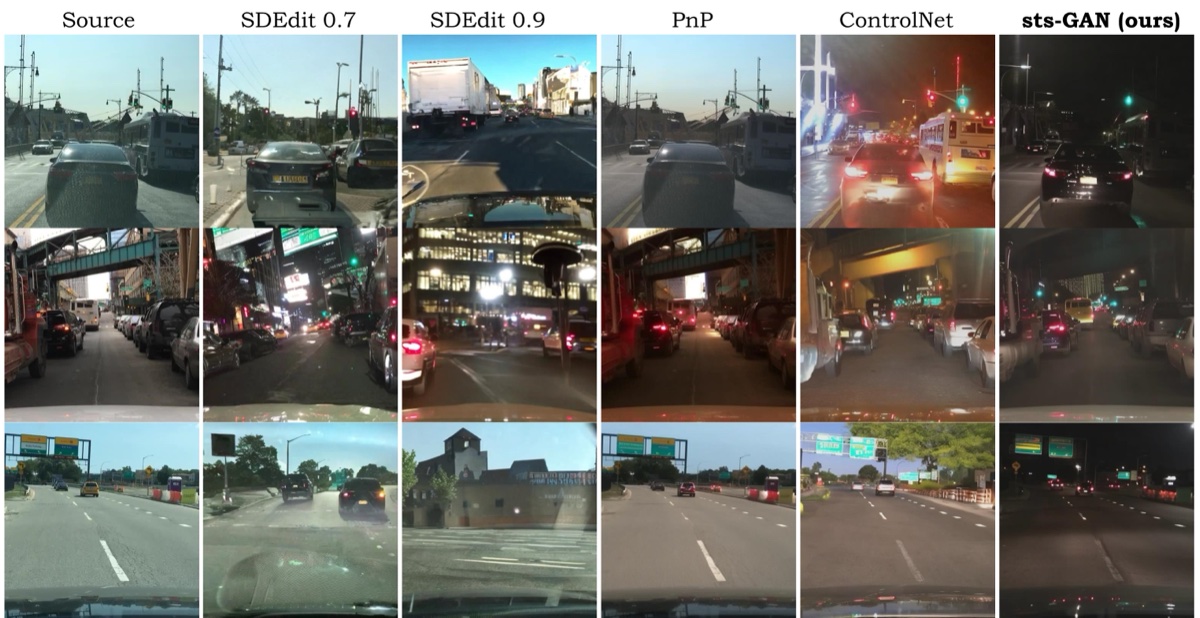}
        \caption{Qualitative comparison for Day-to-Night translation over the \emph{BDD100k} dataset - Diffusion-based baselines}
        \label{subfig:diff}
    \end{subfigure}
    \label{fig:night_comp}
\end{figure}

Quantitative results are presented in \Cref{tab:quant-gan}. Our method achieves the lowest MMD and KID scores and the second lowest FID score. It should be noted that the high SSIM scores achieved by SDEdit and PnP result from their frequent failure to achieve the target domain appearance, as reflected by their low FID, KID, and MMD scores. This phenomenon is explained by the inherent trade-off between achieving the desired target domain appearance and preserving the content from the source image without the cycle-consistency mechanism. 
For example, when increasing the strength parameter of SDEdit above $0.7$, the results become increasingly disconnected from the source image (see \Cref{subfig:diff}). Our model exhibits the best balance between target domain appearance and structure preservation compared to the baselines.

\begin{table}
\centering
\resizebox{\columnwidth}{!}{%
\begin{tabular}{l||ccccc||cccc||c}
\toprule
& \multicolumn{5}{c||}{\textbf{GAN baselines}} & \multicolumn{4}{c||}{\textbf{Diffusion baselines}} & \textbf{StS (ours)} \\
 & CycleGAN & MUNIT & TSIT & AU-GAN & CycleGAN-turbo & SDEdit 0.5 & SDEdit 0.7 & PnP & ControlNet & \\
\midrule
FID $\downarrow$ &
19.908 &
52.152 &
21.315 &
\textcolor{blue}{14.426} &
16.840 &
73.494 &
48.757 &
61.617 &
35.091 &
\textcolor{red}{16.384} \\
MMD $\downarrow$ &
58.395 &
260.081 &
56.484 &
\textcolor{red}{45.970} &
49.845 &
242.001 &
161.666 &
172.808 &
95.171 &
\textcolor{blue}{41.344} \\
KID $\downarrow$ &
4.539 &
12.968 &
4.446 &
\textcolor{red}{3.985} &
4.215 &
12.097 &
9.185 &
9.575 &
6.340 &
\textcolor{blue}{3.718} \\
SSIM $\uparrow$ &
0.469 &
0.308 &
0.3929 &
0.463 &
0.431 &
\textcolor{red}{0.661} &
0.603 &
\textcolor{blue}{0.768} &
0.493 &
0.505 \\
\bottomrule
\end{tabular}
}
\caption{\textbf{Quantitative comparison to other methods.} Day-to-Night translation on the \emph{BDD100k} dataset. For each metric, the best and second-best scores are shown in \textcolor{blue}{blue} and \textcolor{red}{red}, respectively.}
\label{tab:quant-gan}
\end{table}

Qualitatively, \Cref{fig:night_comp} compares our \emph{StS} results to both GAN-based and diffusion-based methods for the Day-to-Night task using the \emph{BDD100k} dataset. Our model achieves the highest level of realism compared to all other methods. The GAN-based methods mostly suffer from the occurrence of artifacts, primarily manifested as random light spots that are uncorrelated with semantically meaningful potential light sources in the image (\eg, car headlights, taillights, streetlights, which are commonly turned off during the day but can be turned on at night). Our model minimizes the occurrence of these artifacts and leverages the powerful semantic understanding of the diffusion model to accurately generate semantics-related target domain effects, such as light sources, light scatters, and reflections (see \Cref{subfig:GAN}). While PnP and SDEdit struggle to balance between output realism and structural preservation, our model excels in both aspects. 

\begin{figure}
    \centering
    \includegraphics[width=0.9\linewidth]{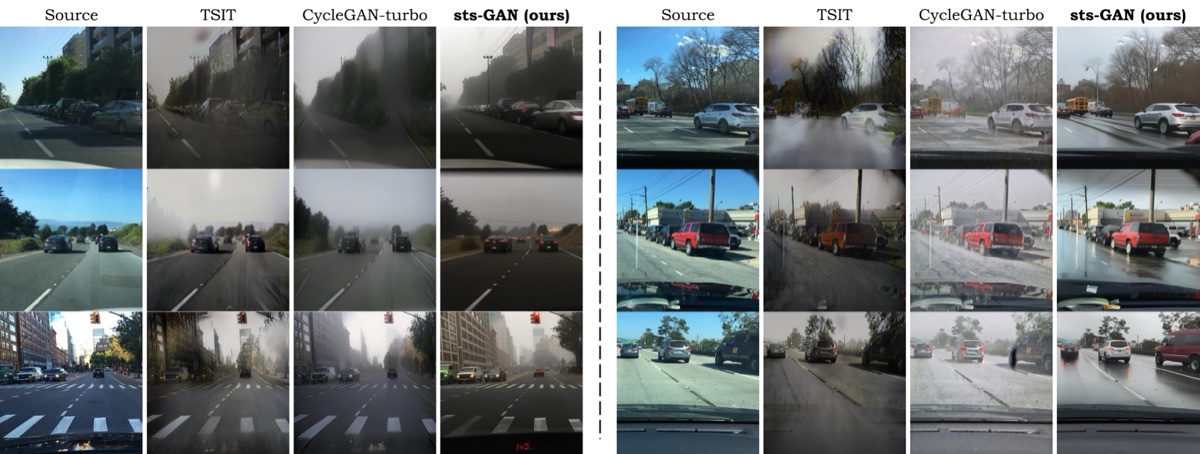}
    \caption{Qualitative comparison of weather translation- \emph{clear2fog} (left) and \emph{clear2rain} (right). See the suppl.~material for additional examples and domains.}
    \label{fig:bdd_domains}
\end{figure}

\Cref{fig:bdd_domains} qualitatively demonstrates our model's performance for weather translations, compared to SoTA GAN-based methods. Specifically, we experimented with Clear-to-Foggy and Clear-to-Rainy translations. Additional results, as well as night-time examples, are provided in the supplementary material.
To train \emph{sts-GAN} for these weather translations, we utilized clear and rainy images from \emph{BDD100k} (both day and night) and foggy images from both the ``light fog'' and ``dense fog'' splits of the \emph{DENSE} dataset~\cite{bijelic2020seeing}. While primarily focused on automotive translations, our model is versatile enough to be generalized for classic object-level edits. Qualitative examples showcasing representative non-automotive applications are provided in \Cref{fig:non_automotive} and are extended in the supplementary material.

\begin{figure}
    \centering
\includegraphics[width=0.99\linewidth]{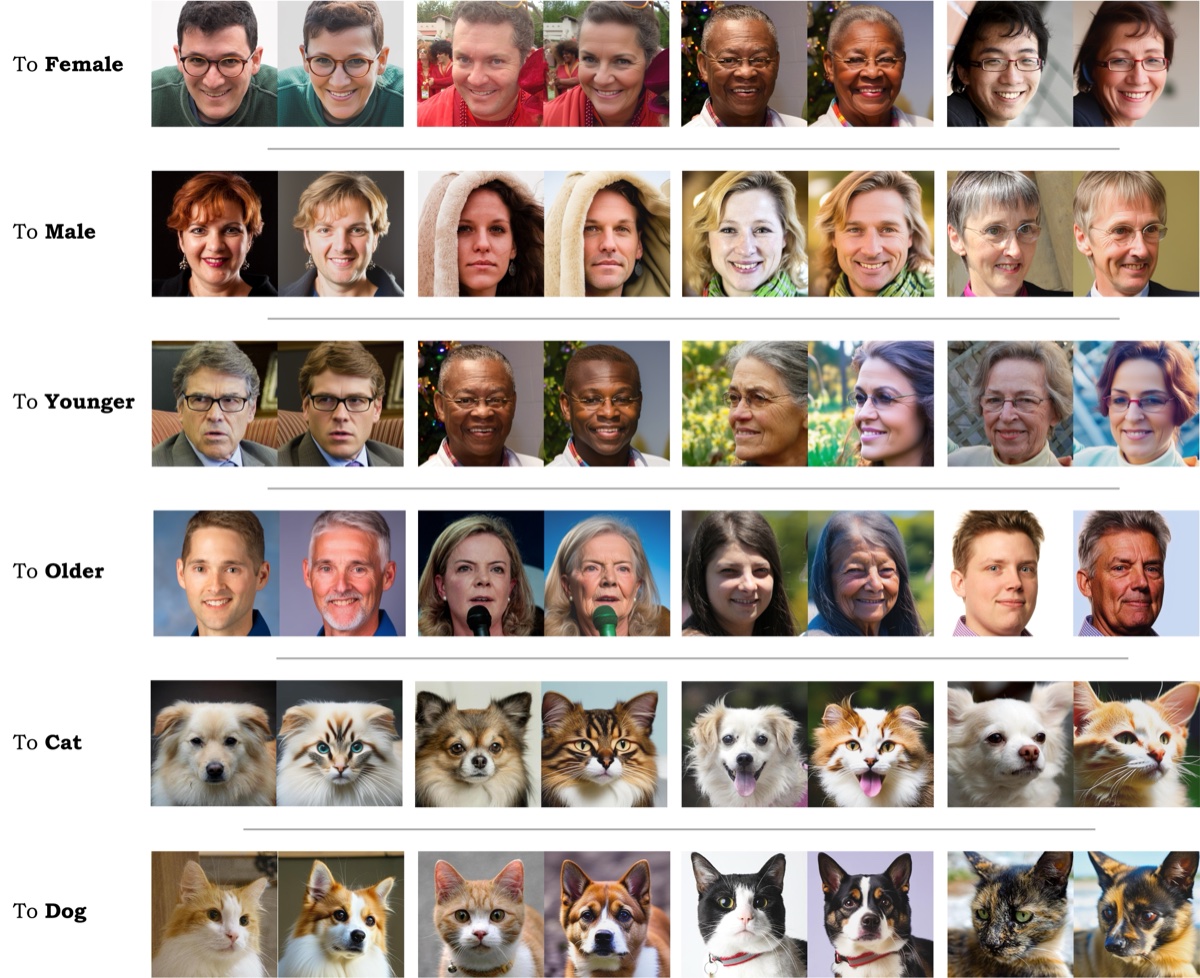}
    \caption{Non-automotive translations on the \emph{FFHQ} \cite{karras2019style} (faces) and \emph{AFHQ} \cite{choi2020stargan} (cats/dogs) datasets. In each pair, the left image is the source and the right image is the translated version.}
    \label{fig:non_automotive}
\end{figure}

\subsection{Ablation Study}
\label{sec:ablation}

\begin{table}
\centering
\resizebox{0.5\columnwidth}{!}{%

    \begin{tabular}{l|c|c|c|c}
          CFG-scale &  FID $\downarrow$&  MMD $\downarrow$& KID $\downarrow$& SSIM $\uparrow$ \\
         \hline \hline
         1.0    &  25.955&  67.404 & 5.364 & 0.549\\
         3.0    & 17.454 &45.540  &3.988 & 0.526\\
         5.0    & 16.384 & 41.344 &3.718& 0.505\\
        \hline
    \end{tabular}}
    \caption{\textbf{CFG Ablation study.} Balance between content preservation and target domain appearance via CFG-scale.}
    \label{tab:ablation_cond}
\end{table}

\begin{table}
\centering
\resizebox{0.8\columnwidth}{!}{%

    \begin{tabular}{l|c|c|c|c}
          Method &  FID $\downarrow$&  MMD $\downarrow$& KID $\downarrow$& SSIM $\uparrow$ \\
         \hline \hline
         ControlNet+RS                &  35.091 (+114\%)&  95.171 (+130\%) & 6.340 (+70\%) & 0.493 (+2\%)\\
         ControlNet+Inv    &49.572 (+202\%) &411.060 (+894\%)  &14.981 (+296\%) & \textbf{0.756 (-49\%)}\\
         ControlNet+RS+ST             &  21.316 (+30\%) & 67.650 (+63\%) & 5.5456 (+49\%) & 0.450 (+11\%)\\
         ControlNet+Inv+ST (\emph{StS}) &  \textbf{16.384}&  \textbf{41.344}& \textbf{3.718}& 0.505\\
        \hline
    \end{tabular}}
    \caption{\textbf{Ablation study - Model Components.} Day-to-Night translation over \emph{BDD100k}.}
    \label{tab:ablation}
\end{table}

We evaluate the contribution of each component in our method through controlled ablations (see \Cref{tab:ablation}). Starting from a baseline ControlNet initialized with a random seed (RS), we incrementally introduce two components: (1) inverted seed initialization (Inv), and (2) seed-to-seed translation via sts-GAN (ST). Our complete model \textbf{StS}, which combines both components, achieves the best results across all metrics, notably improving appearance metrics (FID, KID, MDD) while preserving structure. This confirms that the sts-GAN enables accurate domain transfer while retaining content fidelity. In contrast, using only inverted seeds leads to poor target appearance due to low editability, while using ST over random seeds suffers from structural degradation.

We also ablate the CFG-scale (\Cref{tab:ablation_cond}), finding that $\omega=5.0$ yields the best trade-off between structural preservation and domain consistency. For further analysis and qualitative examples, please refer to the supplementary material.
\section{Conclusions and Future Work}
\label{sec:discuss}

We propose a novel framework for unpaired image-to-image translation that leverages the generative power of pretrained diffusion models (DMs) through seed-space manipulation. Our architecture combines a task-optimized GAN, responsible for unpaired translation in seed-space, with a frozen DM for encoding and image synthesis. This hybrid design allows for structure-preserving translations of complex scenes, outperforming both GAN- and DM-based baselines across various automotive tasks. To mitigate structural drift during sampling, we incorporate ControlNet for spatial guidance. In future work, we aim to replace ControlNet with more general structural control mechanisms. Additionally, we plan to explore whether recent advances in inversion techniques can improve seed quality and structural fidelity through more accurate reconstructions. 

\add{Our method is general and may benefit from integration with emerging DiT-based diffusion backbones (e.g., SD3.x, FLUX). It is also well-suited for distilled models that require fewer sampling steps, where manipulations along the shorter sampling trajectory can be strengthened by a meaningful seed. More broadly, we believe that seed-space manipulations represent a versatile paradigm that extends beyond unpaired image translation, with the potential to open new directions across diverse generative tasks.}

\bibliography{MyBib}

\clearpage
\appendix

\renewcommand{\thesection}{\Alph{section}}
\renewcommand{\thesubsection}{\thesection.\arabic{subsection}}

\setcounter{figure}{0}
\renewcommand{\thefigure}{S\arabic{figure}}
\setcounter{table}{0}
\renewcommand{\thetable}{S\arabic{table}}
\setcounter{equation}{0}
\renewcommand{\theequation}{S\arabic{equation}}

\begin{center}
    \Large\textbf{Supplementary Material}
\end{center}
\vspace{1em}

\section{Deterministic DDIM}

Early denoising diffusion and score-based generative models \cite{ho2020denoising,sohl2015deep,song2019generative} sample seeds from white Gaussian noise and progressively map them to images using a stochastic sampling process.
Denoising Diffusion Implicit Models (DDIM) \cite{song2020denoising} offer a generalization which enables deterministic sampling. In addition to reducing the required number of sampling steps, the DDIM process lends itself to inversion \cite{dhariwal2021diffusion,song2020denoising}, making it possible to map images back to the seed-space. Inversion is crucial for the ability to edit real images using pre-trained diffusion models \cite{wallace2023edict, parmar2023zero, mokady2023null, tumanyan2023plug}. The \emph{deterministic} \ddim{sampling} process that denoises the current sample $x_t$ to yield the next step $x_{t-1}$ can be formulated as:

\begin{equation}
\label{eq:ddimSamp}
x_{t-1} = \sqrt{\alpha_{t-1}}\cdot \pred{x}{0} 
+ \sqrt{1-\alpha_{t-1}} \cdot \epsilon_{\theta}^{t}({x_t})
\end{equation}

where $\pred{x}{0}$ is a prediction of the final denoised sample $x_0$ from $x_t$, given by:
\begin{equation}
\label{eq:predx0}
\pred{x}{0} = \frac{x_t -\sqrt{1-\alpha_t}\cdot \epsilon_{\theta}^{t}({x_t})}{\sqrt{\alpha_t}}.
\end{equation}

Here $\alpha_{t-1}, \alpha_t$ are the per-timestep diffusion schedule hyperparameters, and 
$\epsilon_{\theta}^{t}$ is the noise prediction U-net, parameterized by $\theta$.

The reverse process, referred to as \ddim{inversion}, is formulated as follows (at the limit of decreasing step size):

\begin{equation}
\label{eq:ddimInv}
x_{t+1} = \sqrt{\alpha_{t+1}}\cdot \pred{x}{0} + \sqrt{1-\alpha_{t+1}} \cdot \epsilon_{\theta}^{t}({x_t})
\end{equation}

\section{Implementation Details}

In this section, we provide detailed implementation and training information regarding the different models that were trained during this work.

\subsection{Seed-Space Classifier}

We use a uniform \emph{ResNet18}-based classifier for all classification tasks presented in Table 1 in the main text. For the tasks applied within the seed-space, we adjust the first layer of the classifier to 4-channeled input to fit the dimensionality of Stable Diffusion's latent representation. We split the training data to $80\%$ for training and $20\%$ for validation. We trained all models to a maximum of 80 epochs over the training set, and chose the best accuracy over the validation. We use the Adam optimizer \cite{kingma2014adam} with lr=0.001 for all tasks. Task-specific details are provided below:

\begin{itemize}
    \item \textbf{day\textbackslash night:} We trained over \emph{BDD100k} ``daytime'' and ``night''  splits. For the seed-space version, we first center-cropped each sample to $512\times512$, then inverted them to the seed space.
    \item \textbf{cat\textbackslash dog:} We trained over the ``cat'' and ``dog'' splits of the \emph{AFHQ} \cite{choi2020stargan} dataset without additional preproccessing.
    \item \textbf{older\textbackslash younger:} We used the provided metadata of the \emph{FFHQ} \cite{karras2019style} dataset and chose samples tagged as 55+ years old as the ``older'' split and those tagged in the range of 17-40 years old as the ``younger'' split. We used the $512\times512$ version of the dataset.
\end{itemize}

\subsection{Finetuning SD for Automotive Dataset}
\label{suppsubsec:finetuning}
The pre-trained version of Stable Diffusion (SD) 2.1 performs poorly on realistic driving datasets. As a result, we fine-tune SD 2.1 using the \emph{BDD100k} training set. We automatically generate the textual conditions using information provided in the dataset's metadata logs regarding Weather and Time-Of-Day. The resulting prompts have the form:
\begin{quote}
``A \textit{\textcolor{magenta}{*Weather*}} \textit{\textcolor{blue}{*Time-Of-Day*}}'' 
\end{quote}

The various choices available in the metadata logs of \emph{BDD100k} for individual attributes are delineated in \Cref{tab:attr}. It should be noted that all images featuring an ``undefined'' label for any attribute have been excluded from the training set. ``dawn/dusk'' images were also excluded due to low amount of samples and unclear thresholds between ``dawn/dusk'' and ``daytime/night''.

\begin{table}
    \centering
    \begin{tabular}{c|p{5cm}}
         \hline
         \textbf{Weather}&  
         rainy, snowy, clear, overcast, undefined, partly cloudy, foggy\\
         \hline
         \textbf{Time-Of-Day}&  daytime, night, dawn/dusk, undefined\\
         \hline
    \end{tabular}
    \caption{Attributes and corresponding options provided in BDD100k metadata logs.}
    \label{tab:attr}
\end{table}

\begin{figure}
	\centering
	\subfigure[Pre-trained model]{\label{fig:subfigA}\includegraphics[width=0.6\linewidth]{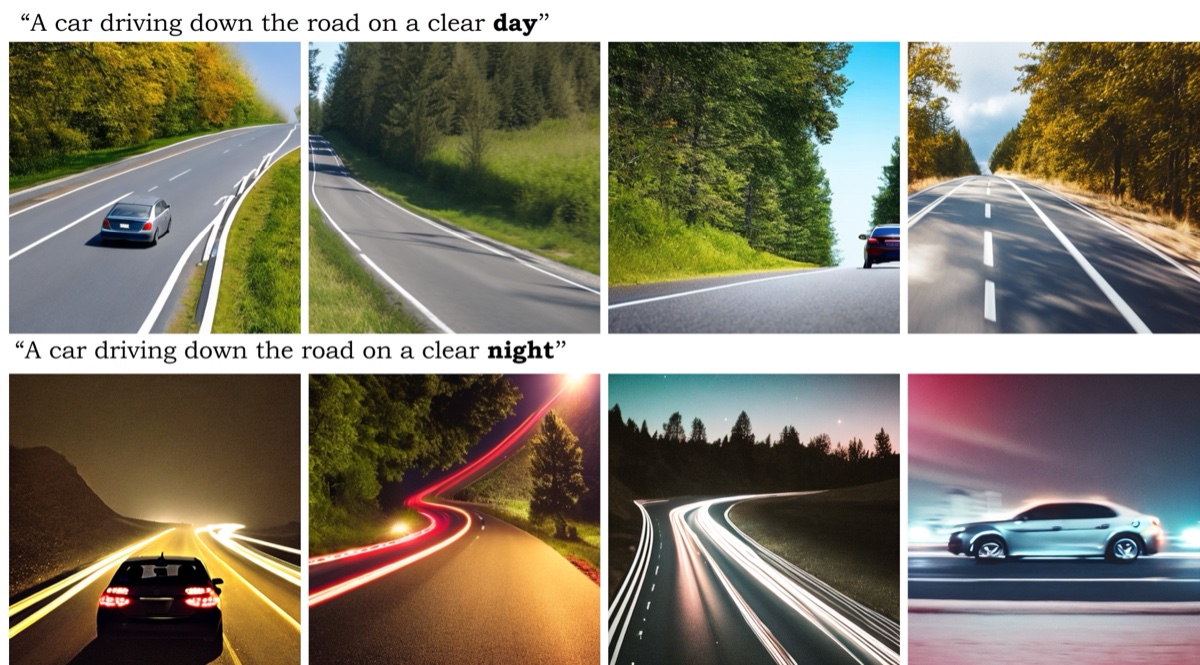}} \\
	\vspace{1em}
	\subfigure[Fine-Tuned model]{\label{fig:subfigB}\includegraphics[width=0.6\linewidth]{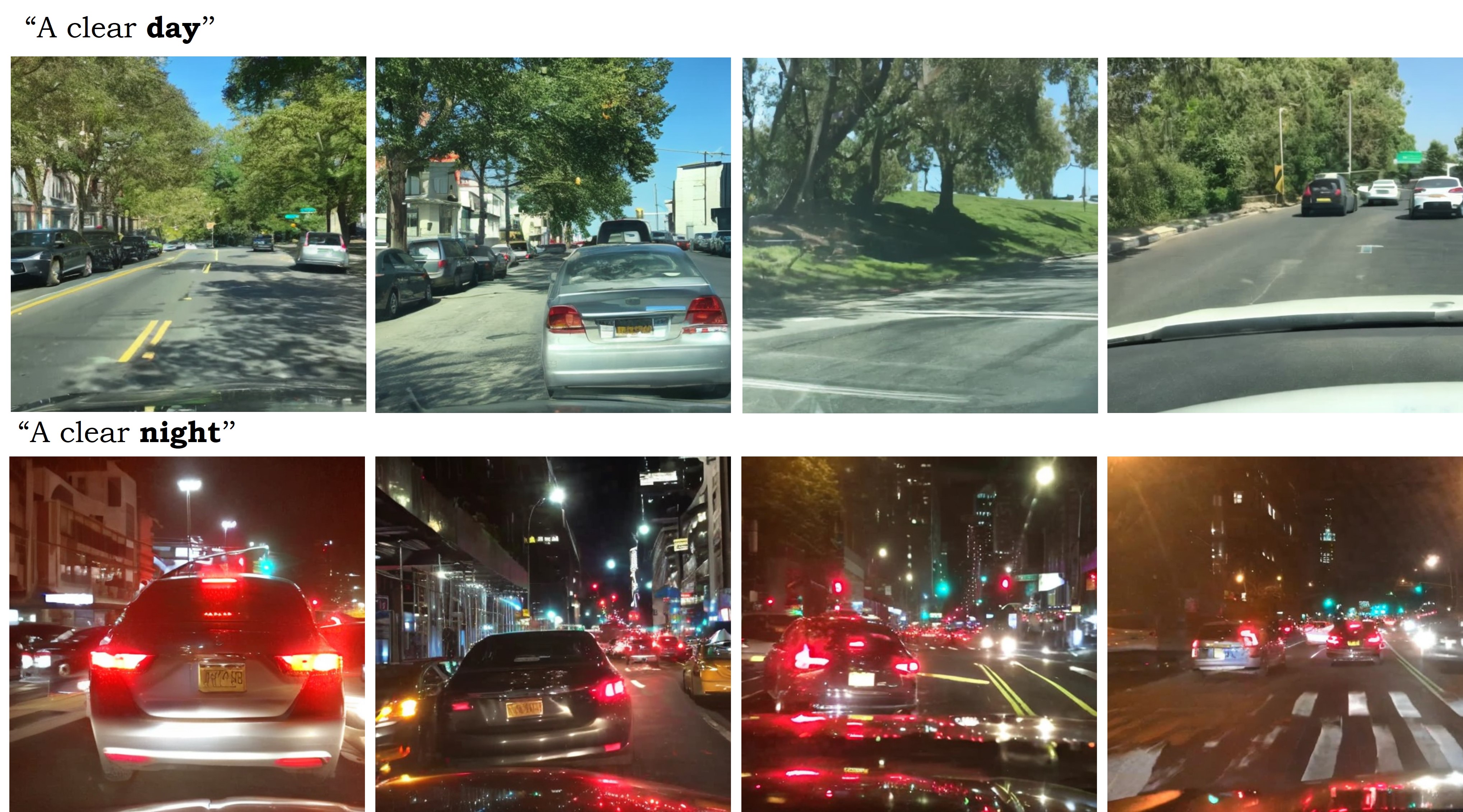}}
	\caption{Fine-tuning SD 2.1 for automotive images using the BDD100k dataset: (a) before, and (b) after.}
	\label{fig:without_control}
\end{figure}

Some synthetic images before and after fine tuning are illustrated in Figure \ref{fig:without_control}.

\begin{figure}
	\centering
	\subfigure[Pre-trained model]{\label{fig:subfigA}\includegraphics[width=0.6\linewidth]{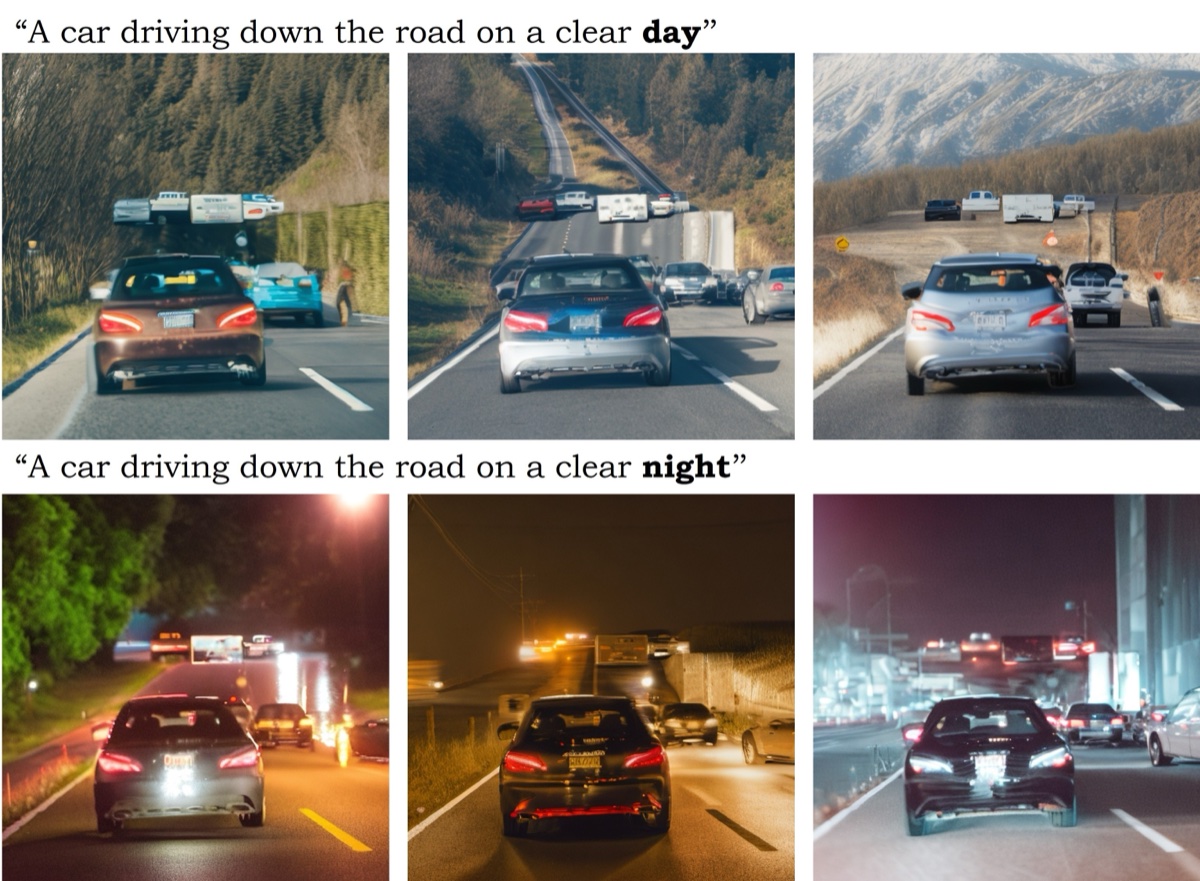}} \\
	\vspace{1em}
	\subfigure[Fine-Tuned model]{\label{fig:subfigB}\includegraphics[width=0.6\linewidth]{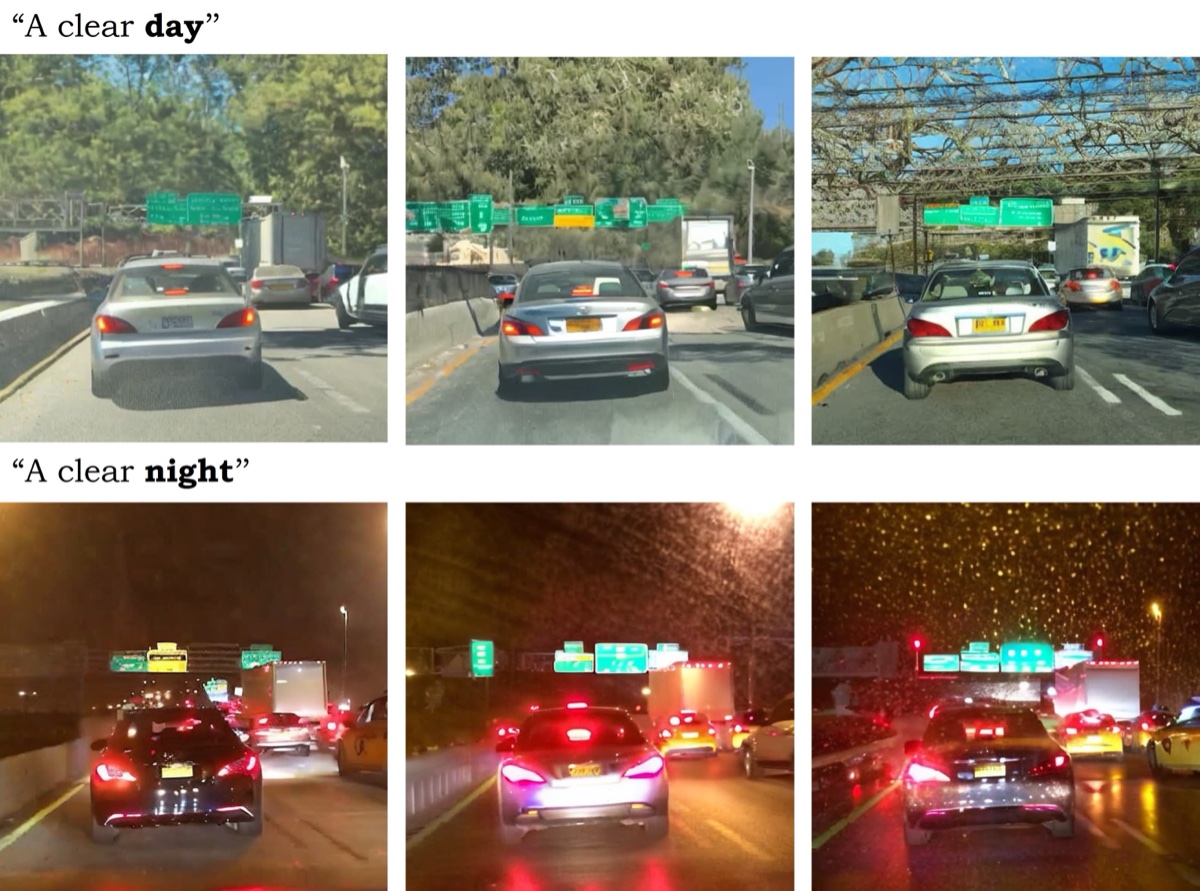}}
	\caption{Trained ControlNet over fine-tuned SD 2.1 for automotive, vs. pretrained ControlNet from \cite{Thibaud2023controlnet}.}
	\label{fig:with_control}
\end{figure}

We train a ControlNet over our fine-tuned SD using the same dataset. We utilize a Canny-like spatial control, derived by applying a Canny edge detector over a segmentation mask obtained using the publicly available version of the Segment-Anything Model \cite{kirillov2023segany} (SAM). This approach ensures that only the boundaries of each object and sub-object are considered. Through experimentation, we found this spatial control to be superior to using Canny directly with different thresholds or a direct SAM mask. Some controlled synthetic images before and after fine tuning are illustrated in Figure \ref{fig:with_control}. 

\begin{figure}
    \centering
  \includegraphics[width=0.6\linewidth]{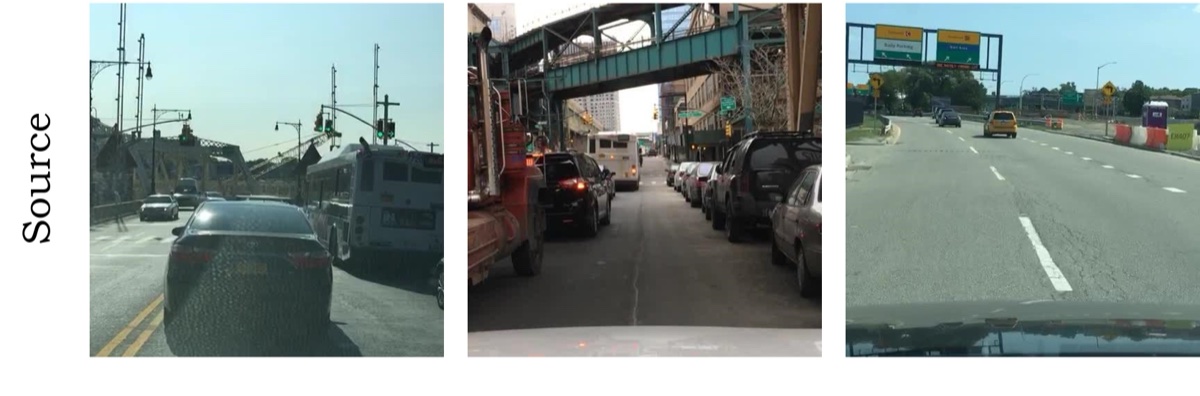}
  \includegraphics[width=0.6\linewidth]{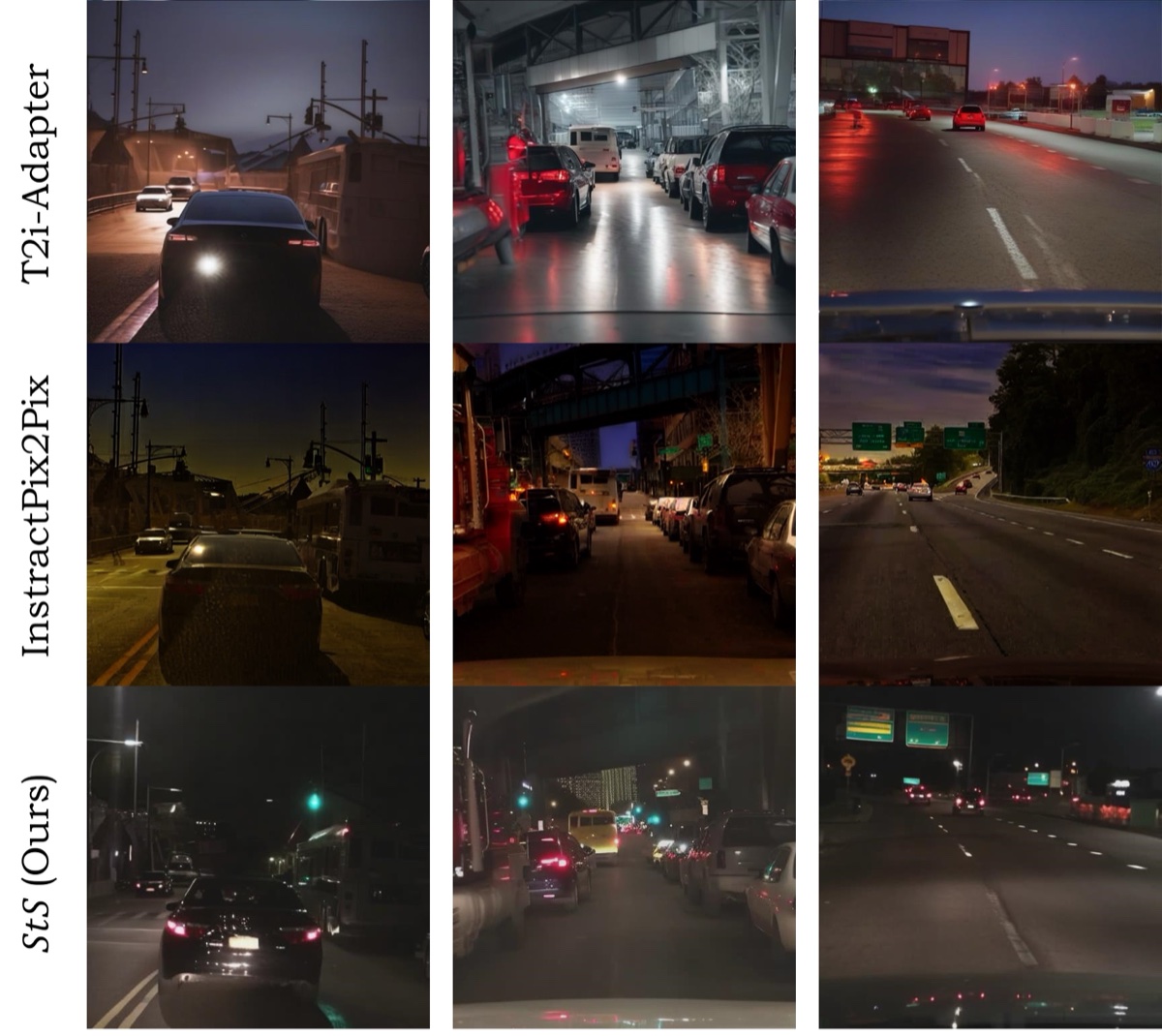}
    \caption{A qualitative comparison between our method and existing diffusion-based methods for day-to-night translation. This figure is an extension of Figure 7b in the main text, that 
    includes two additional diffusion-based methods (T2I-Adapters \cite{mou2023t2iadapter} and InstructPix2Pix \cite{brooks2023instructpix2pix}), each of which involves a specific training process and could not utilize our fine-tuned U-net.
    }
    \label{fig:notune}
\end{figure}

\Cref{fig:notune} shows a qualitative comparison between our method and additional diffusion-based baselines for the day-to-night translation task. Specifically, it extends the comparison shown in Figure 7b in the main text by including two additional methods, T2I-Adapters \cite{mou2023t2iadapter} and InstructPix2Pix \cite{brooks2023instructpix2pix}. These methods involve a training process specific to them, and could not utilize our fine-tuned U-net.

\begin{figure}
    \centering
  \includegraphics[width=0.8\linewidth]{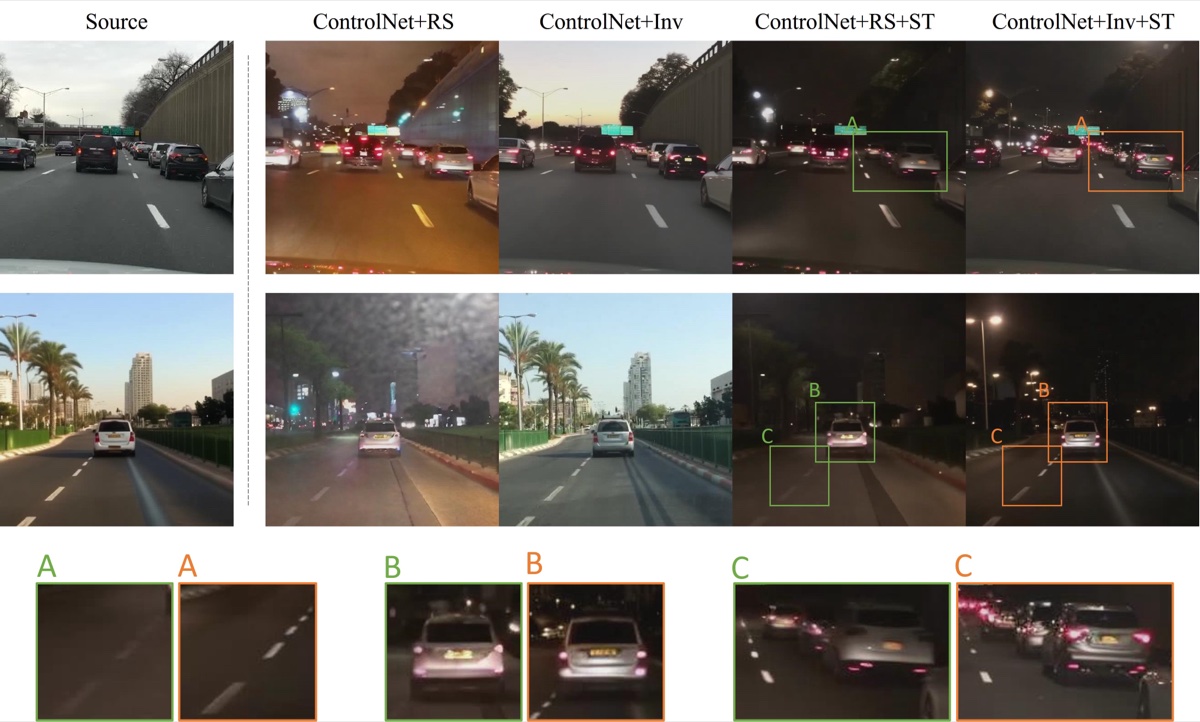}
    \caption{\textbf{Qualitative Ablation Study.} Day-to-Night translation over \emph{BDD100k}. All samples were generated using the prompt ``A clear night''.}
    \label{fig:ablation}
\end{figure}

\begin{table}
\centering
\resizebox{0.5\columnwidth}{!}{%

    \begin{tabular}{l|c|c|c|c}
          CFG-scale &  FID $\downarrow$&  MMD $\downarrow$& KID $\downarrow$& SSIM $\uparrow$ \\
         \hline \hline
         1.0    &  25.955&  67.404 & 5.364 & 0.549\\
         3.0    & 17.454 &45.540  &3.988 & 0.526\\
         5.0    & 16.384 & 41.344 &3.718& 0.505\\
        \hline
    \end{tabular}}
    
    \caption{\textbf{CFG Ablation study.} Balance between content preservation and target domain appearance via CFG-scale.}
    \label{tab:ablation_cond}
\end{table}

\section{Ablation Study}
\label{sec:ablation}

\begin{table}
\centering
\resizebox{0.8\columnwidth}{!}{%

    \begin{tabular}{l|c|c|c|c}
          Method &  FID $\downarrow$&  MMD $\downarrow$& KID $\downarrow$& SSIM $\uparrow$ \\
         \hline \hline
         ControlNet+RS                &  35.091 (+114\%)&  95.171 (+130\%) & 6.340 (+70\%) & 0.493 (+2\%)\\
         ControlNet+Inv    &49.572 (+202\%) &411.060 (+894\%)  &14.981 (+296\%) & \textbf{0.756 (-49\%)}\\
         ControlNet+RS+ST             &  21.316 (+30\%) & 67.650 (+63\%) & 5.5456 (+49\%) & 0.450 (+11\%)\\
         ControlNet+Inv+ST (\emph{StS}) &  \textbf{16.384}&  \textbf{41.344}& \textbf{3.718}& 0.505\\
        \hline
    \end{tabular}}
    \caption{\textbf{Ablation study - Model Components.} Day-to-Night translation over \emph{BDD100k}.}
    \label{tab:ablation}
\end{table}

We examine the effectiveness of the different components of our approach by incorporating them one at a time (see \Cref{tab:ablation}). On top of a pure off-the-shelf ControlNet initiated with a random seed (RS) $z_T \sim \mathcal{N}(0,I)$ (\emph{ControlNet+RS}), we add two components that make up the complete Seed Translation (ST) block: (1) initiation with a meaningful seed obtained by \ddim{inversion} (\emph{ControlNet+Inv}) and (2) using the \emph{sts-GAN} in the seed space for the seed translation (\emph{ControlNet+Inv+ST}). We also demonstrate the contribution of the \emph{sts-GAN} block applied over a random seed, rather than an inverted one (\emph{ControlNet+RS+ST}). The contribution of each component is reported in \Cref{tab:ablation} and qualitatively illustrated in \Cref{fig:ablation}. In all configurations we used the whole \emph{BDD100k} test set with 20 DDIM steps and CFG-scale $\omega=5.0$.

Notably, initiating the ControlNet sampling process with the inverted seed (\emph{ControlNet+Inv}) imposes an overly rigid constraint, significantly reducing editability during sampling. As proposed in the introduction, the semantic information encoded within the inverted seed includes daytime-related attributes, which conflict with the desired appearance indicated by the textual guidance, resulting in low editability. This challenge is reflected in the combination of a very high SSIM score alongside poor appearance metrics.

In the second configuration (\emph{ControlNet+RS+ST}) we randomly sampled an initial seed, then translated it into a target-related one using our \emph{sts-GAN}. The translated seed is subsequently sampled using ControlNet.
Here, the seed that initiates the sampling process unconditionally possesses attributes of the target domain, hence results with more accurate appearance of the output image in the target domain, which is reflected in better appearance measures compared to \emph{ControlNet+RS}. Yet, the generated images tend to over-blurriness, reflected by a lower SSIM score. 

Combining the ST block with an initially meaningful inverted seed (\emph{ControlNet+Inv+ST} (\emph{StS})) yields the desired combination of best appearance and high fidelity to the structure. The details are sharper and more accurate, compared to the \emph{ControlNet+RS+ST} configuration (as illustrated by crops \emph{A, B} and \emph{C} in \Cref{fig:ablation}), which lead to an improvement of ~11\% in the SSIM score. Nonetheless, the images generated by our \emph{StS} model commonly exhibit more accurate and semantically meaningful \emph{local} target-related effects. When a random seed is being translated to a target-domain-related seed using \emph{sts-GAN}, the translated seed encodes information about the global appearance of the target domain but lacks details about the local semantics of the source image. Consequently, local, semantics-related effects are better generated using the translated-inverted seed than a random-translated one. This phenomenon is evident in features such as head/tail lights, street lights, reflections, etc., and is quantitatively demonstrated by superior performance in target-appearance metrics. This ablation provides insight into both the contribution of the \emph{sts-GAN} and what it has actually learned.

As mentioned, the spatial control stabilizes the loss of details caused by the CFG mechanism. While contributing to the target-domain appearance of the output (see Figure 3 in the main text), it somewhat reduces the structural preservation, as quantitatively evaluated in \Cref{tab:ablation_cond}.

\section{Additional DM-based baselines}

As detailed in the related work in the main text, we chose to evaluate our method against DM-based methods that impose global constraints on the source image and therefore allow global edits while maintaining adherence to the source image. In this section, we demonstrate the limitations of locally constrained and non-constrained methods in such edits.

\begin{figure}
	\centering
	\subfigure[outdoor scenes, from the \emph{COCO2017} dataset]{\label{fig:baselines_coco}\includegraphics[width=0.9\linewidth]{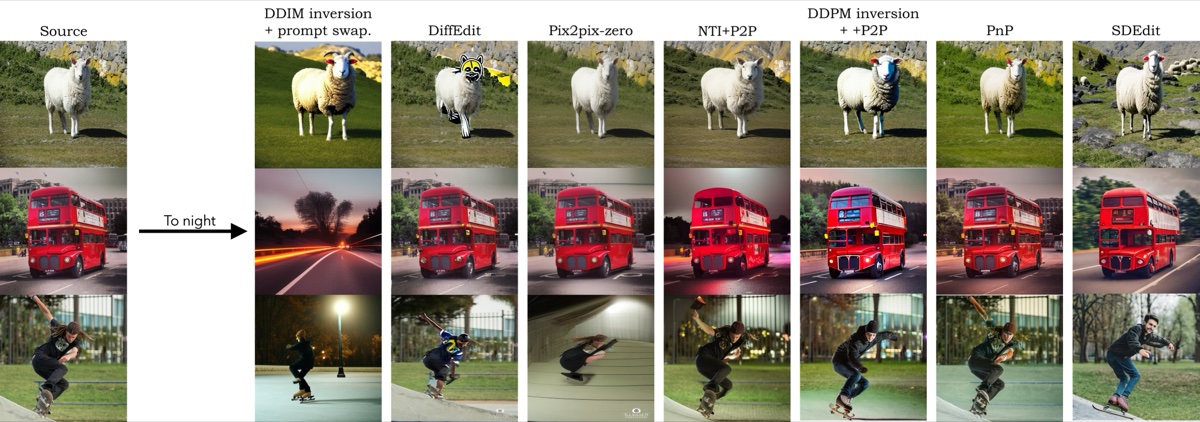}} \\
	\vspace{1em}
	\subfigure[automotive scenes, from the \emph{BDD100K} dataset]{\label{fig:baselines_bdd}\includegraphics[width=0.9\linewidth]{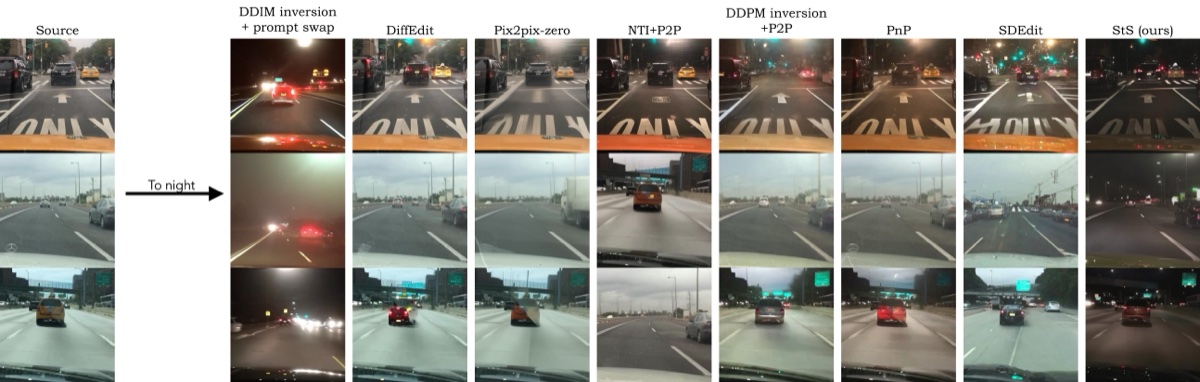}}
	\caption{Comparison of different DM-based methods on the global translation of day-to-night}
	\label{fig:baselines}
\end{figure}

Locally constrained methods, such as Prompt-to-Prompt (P2P) \cite{hertz2022prompt}, Pix2Pix-Zero \cite{parmar2023zero}, and DiffEdit \cite{couairon2022diffedit}, use local spatial constraints in the form of attention maps or dynamically generated masks to focus edits on specific regions of the image while leaving the rest mostly unchanged. \Cref{fig:baselines_coco} demonstrates global edits (day-to-night) on outdoor samples from the \emph{COCO2017} dataset \cite{lin2014microsoft} using non-constrained DDIM-inversion with prompt swapping, locally constrained methods including P2P \cite{hertz2022prompt} with NTI \cite{mokady2023null} and DDPM-Inversion \cite{huberman2024edit}, pix2pix-zero \cite{parmar2023zero}and DiffEdit \cite{couairon2022diffedit}  as well as globally-constrained methods: PnP \cite{tumanyan2023plug} and SDEdit \cite{meng2021sdedit}. All methods use SD2.1 with the default hyperparameters provided by their respective authors. Notably, non-constrained methods (DDIM inversion + prompt swap) fail to preserve the details of the source image. Both Locally and globally constrained methods struggle to achieve global edits while maintaining source image details. In most cases, the target domain (night) is almost not reflected in the translated images. 

\Cref{fig:baselines_bdd} demonstrated the performance of the aformentioned methods on automotive scenes from the \emph{BDD100k} dataset using the fine-tuned Unet (see \Cref{suppsubsec:finetuning}.)

\section{Additional Examples}

\begin{figure}
	\centering
	\subfigure[Clear day to foggy day]{\label{fig:subfigA}\includegraphics[width=\linewidth]{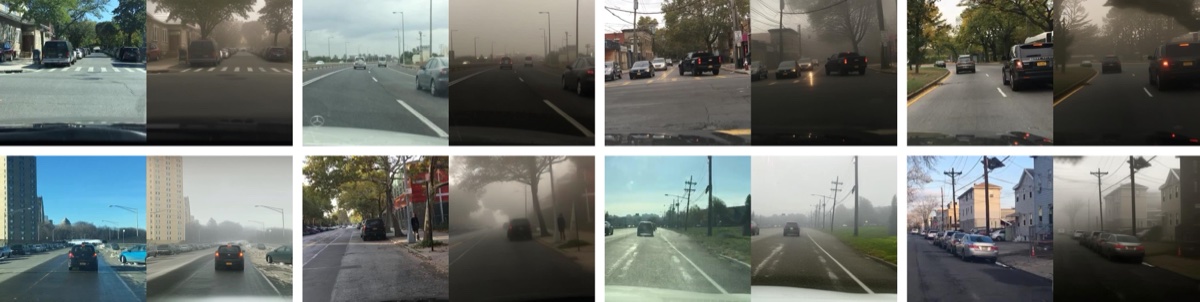}} \\
	\vspace{0.5em}
	\subfigure[Clear night to foggy night]{\label{fig:subfigB}\includegraphics[width=\linewidth]{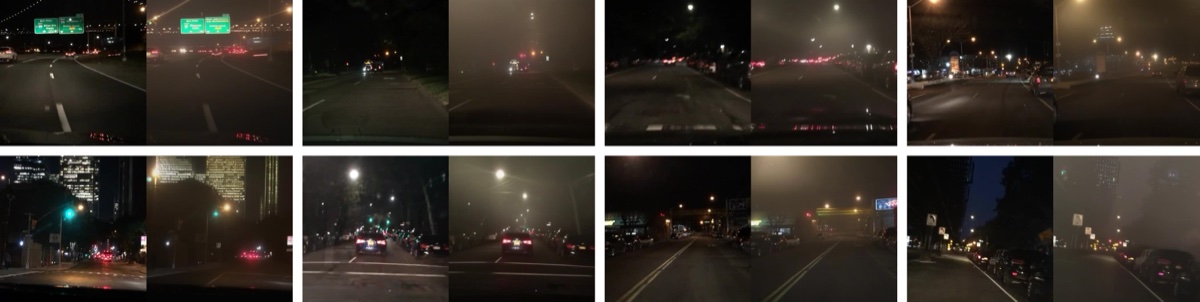}} \\
	\vspace{0.5em}
	\subfigure[Clear day to rainy day]{\label{fig:subfigC}\includegraphics[width=\linewidth]{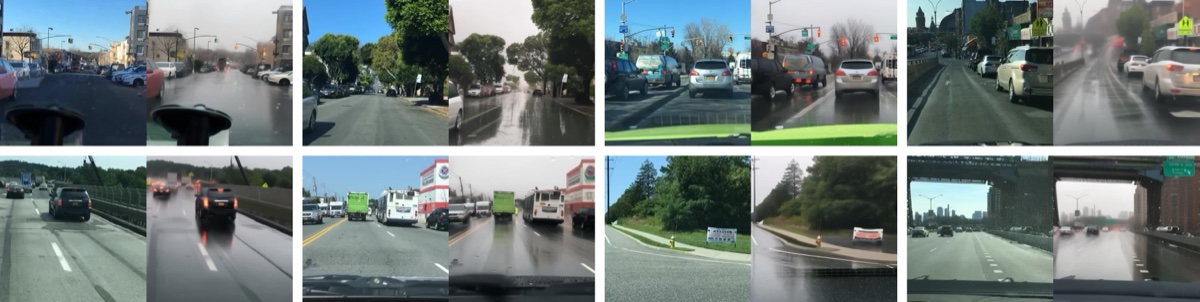}} \\
	\vspace{0.5em}
	\subfigure[Clear night to rainy night]{\label{fig:subfigD}\includegraphics[width=\linewidth]{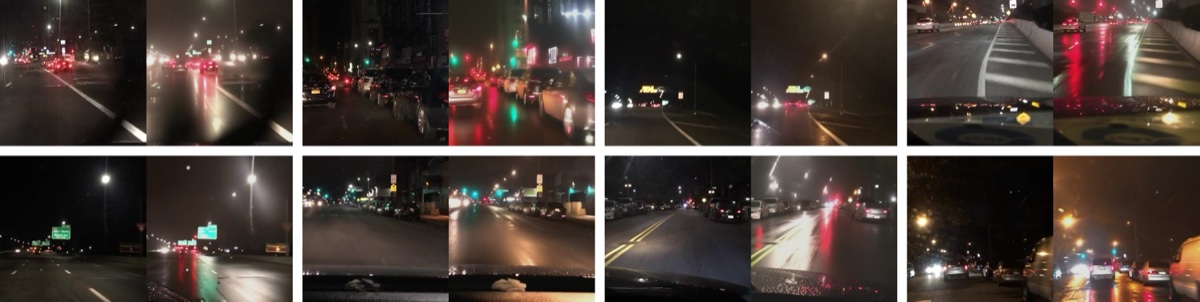}}
	\caption{Weather Translation over the \emph{BDD100k} Dataset. In each pair, the left image is the source and the right is the translated.}
	\label{fig:weather}
\end{figure}

\Cref{fig:weather} presents more examples of different weather translation performed over the \emph{BDD100k} dataset.

\section{Non-Automotive Applications}

Though focused on automotive-related translations, our model is also suitable for additional diverse translation tasks on other datasets. In this section we provide some examples of non-automotive translations using our proposed \emph{sts-GAN}, followed by a short discussion of possible limitations.

\subsection{Non-automotive examples}

\begin{figure}
	\centering
	\subfigure[Male-to-female]{\label{fig:subfigA}\includegraphics[width=\linewidth]{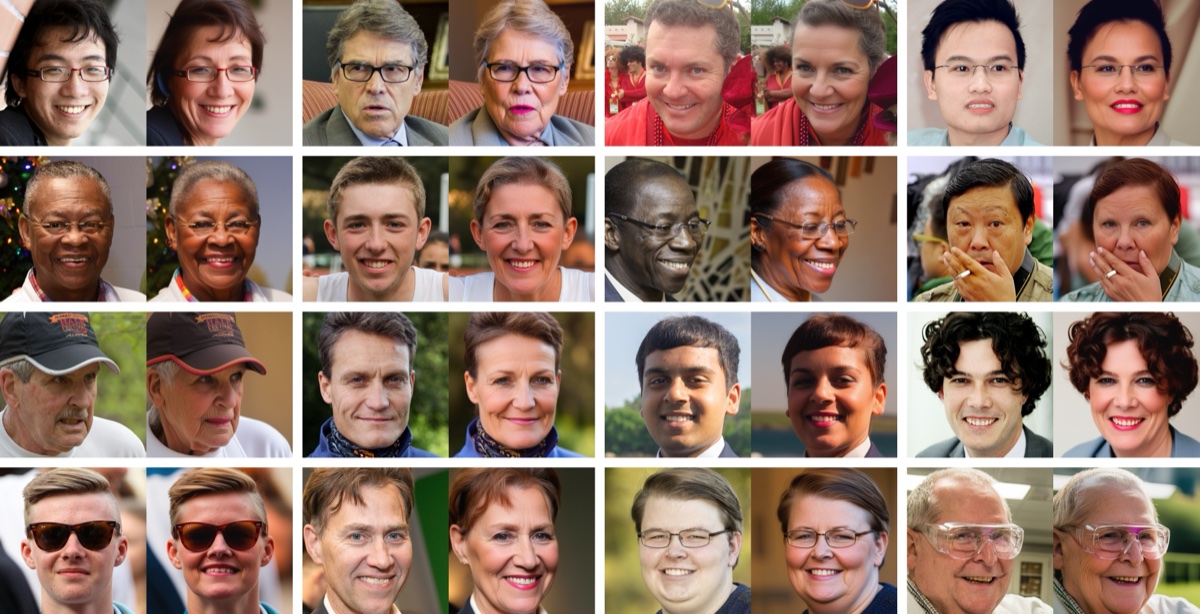}} \\
	\vspace{0.5em}
	\subfigure[Female-to-male]{\label{fig:subfigB}\includegraphics[width=\linewidth]{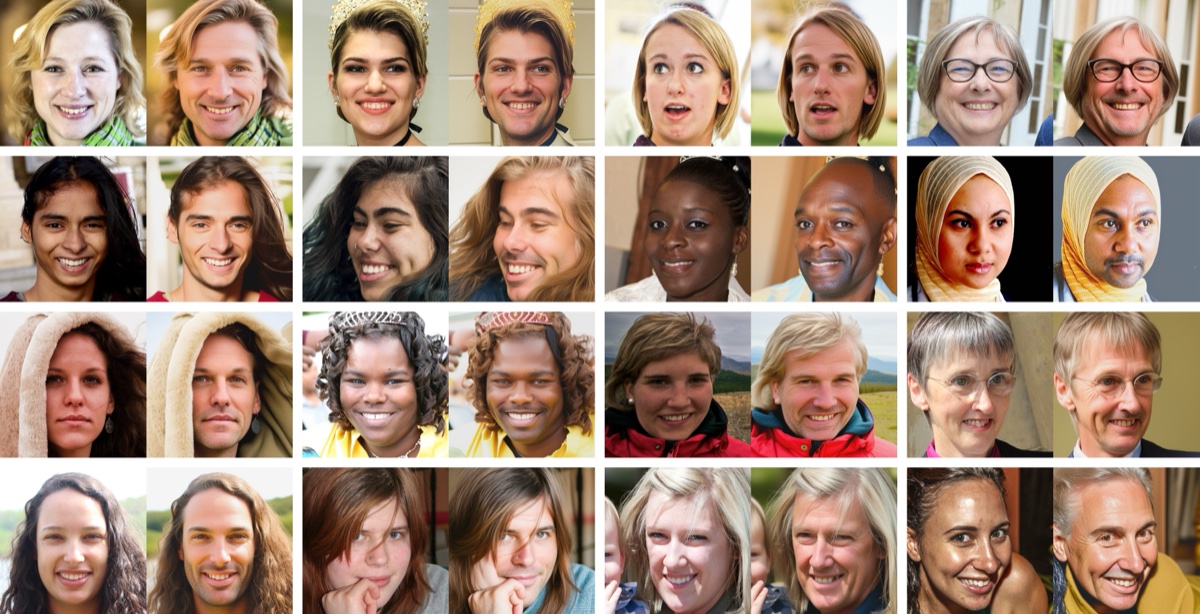}}
	\caption{\emph{Male}$\leftrightarrow$\emph{Female} translation over the \emph{FFHQ} dataset. In each pair, the left image is the source and the right is the translated.}
	\label{fig:gender}
\end{figure}

\begin{figure}
	\centering
	\subfigure[To younger]{\label{fig:subfigA}\includegraphics[width=\linewidth]{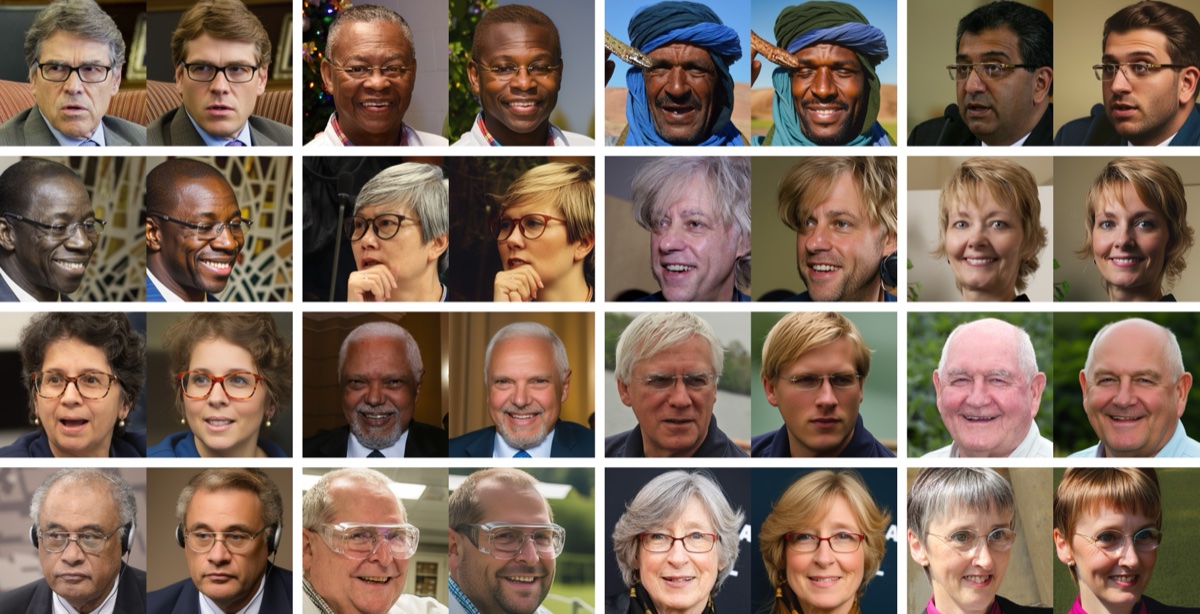}} \\
	\vspace{0.5em}
	\subfigure[To older]{\label{fig:subfigB}\includegraphics[width=\linewidth]{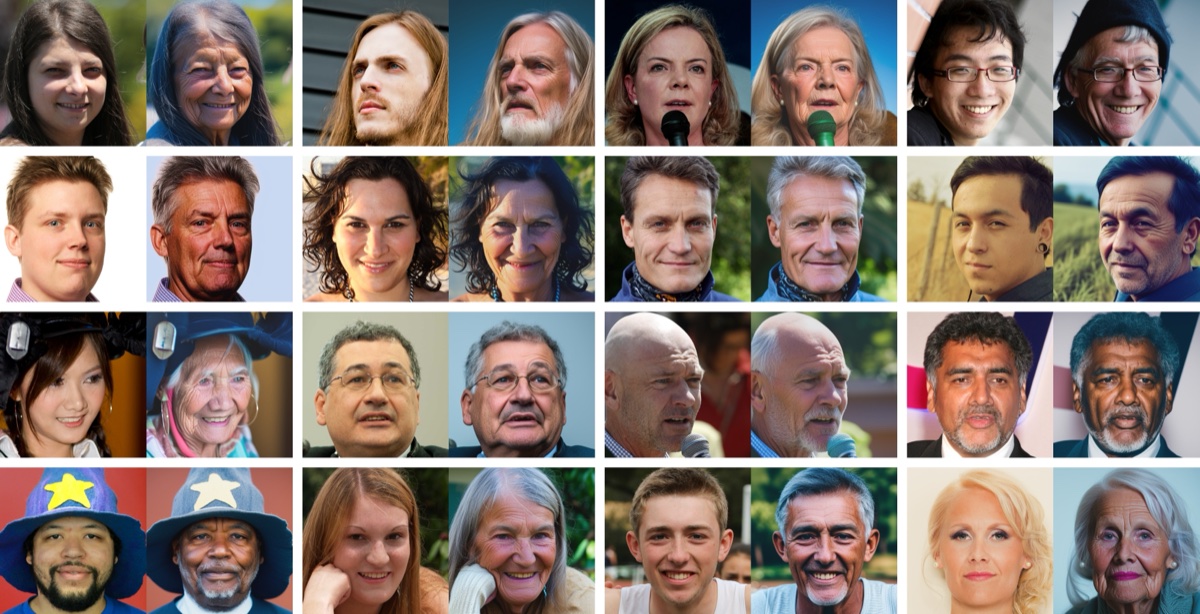}}
	\caption{\emph{Younger}$\leftrightarrow$\emph{Older} translation over the \emph{FFHQ} dataset. In each pair, the left image is the source and the right is the translated.}
	\label{fig:age}
\end{figure}

\begin{figure}
	\centering
	\subfigure[Dog-to-cat]{\label{fig:subfigA}\includegraphics[width=\linewidth]{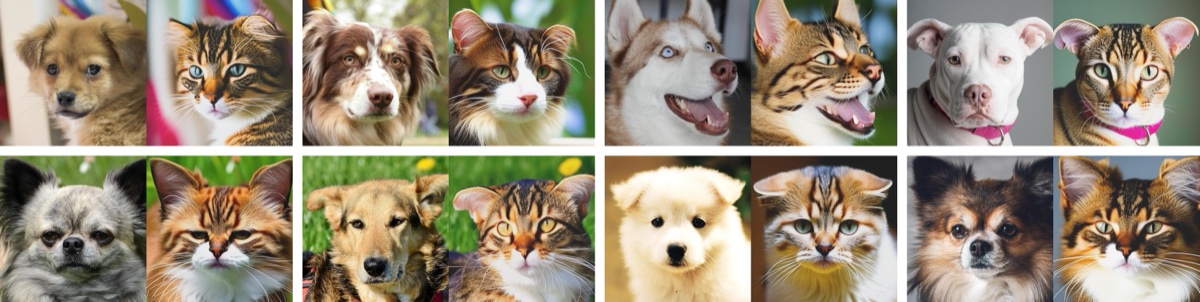}} \\
	\vspace{0.5em}
	\subfigure[Cat-to-Dog]{\label{fig:subfigB}\includegraphics[width=\linewidth]{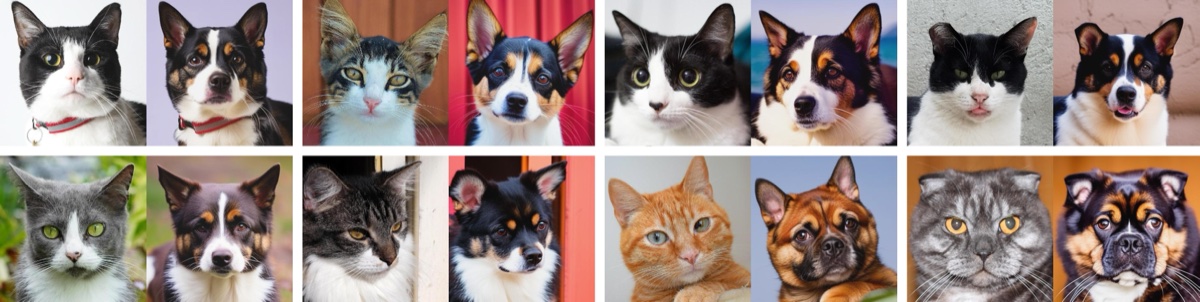}}
	\caption{\emph{Cat}$\leftrightarrow$\emph{Dog} translation over the \emph{AFHQ} dataset. In each pair, the left image is the source and the right is the translated.}
	\label{fig:dog_cat}
\end{figure}

\Cref{fig:gender}, \Cref{fig:age} and \Cref{fig:dog_cat} demonstrate our model's performance in the more common fields of face editing (gender swap and aging, respectively) and object-swap (cat$\leftrightarrow$dog). For the faces and cat/dog tasks we used the \emph{FFHQ} \cite{karras2019style} and \emph{AFHQ} \cite{choi2020stargan} datasets, respectively. For both datasets we used the publicly available pretrained versions of SD2.1-base \cite{rombach2022high} with the pretrained version on ControlNet provided by \cite{Thibaud2023controlnet} without any additional finetuning. \emph{Sts-GAN} was trained using the same procedure as in the automotive case.

\begin{figure}
	\centering
	\subfigure[Female-to-Male]{\label{fig:subfigA1}\includegraphics[width=0.23\textwidth]{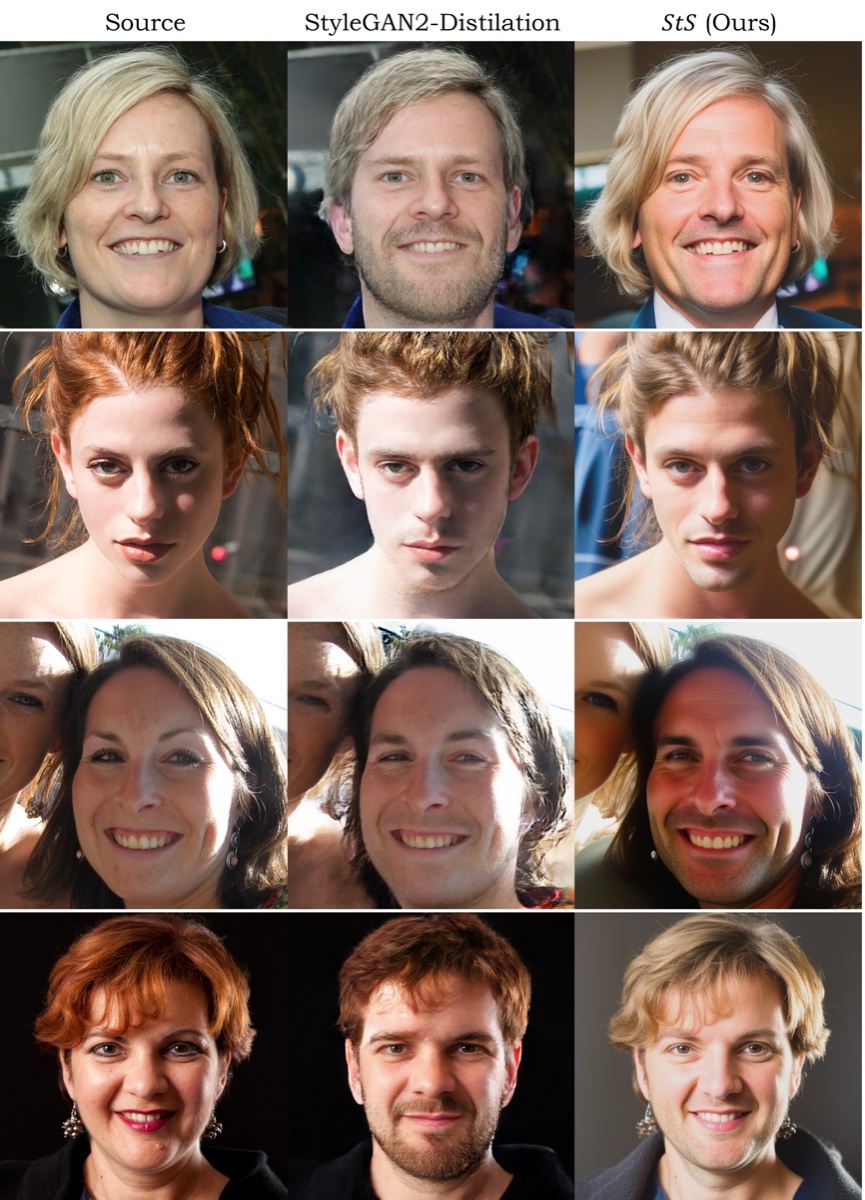}} \hfill
	\subfigure[Male-to-Female]{\label{fig:subfigB1}\includegraphics[width=0.23\textwidth]{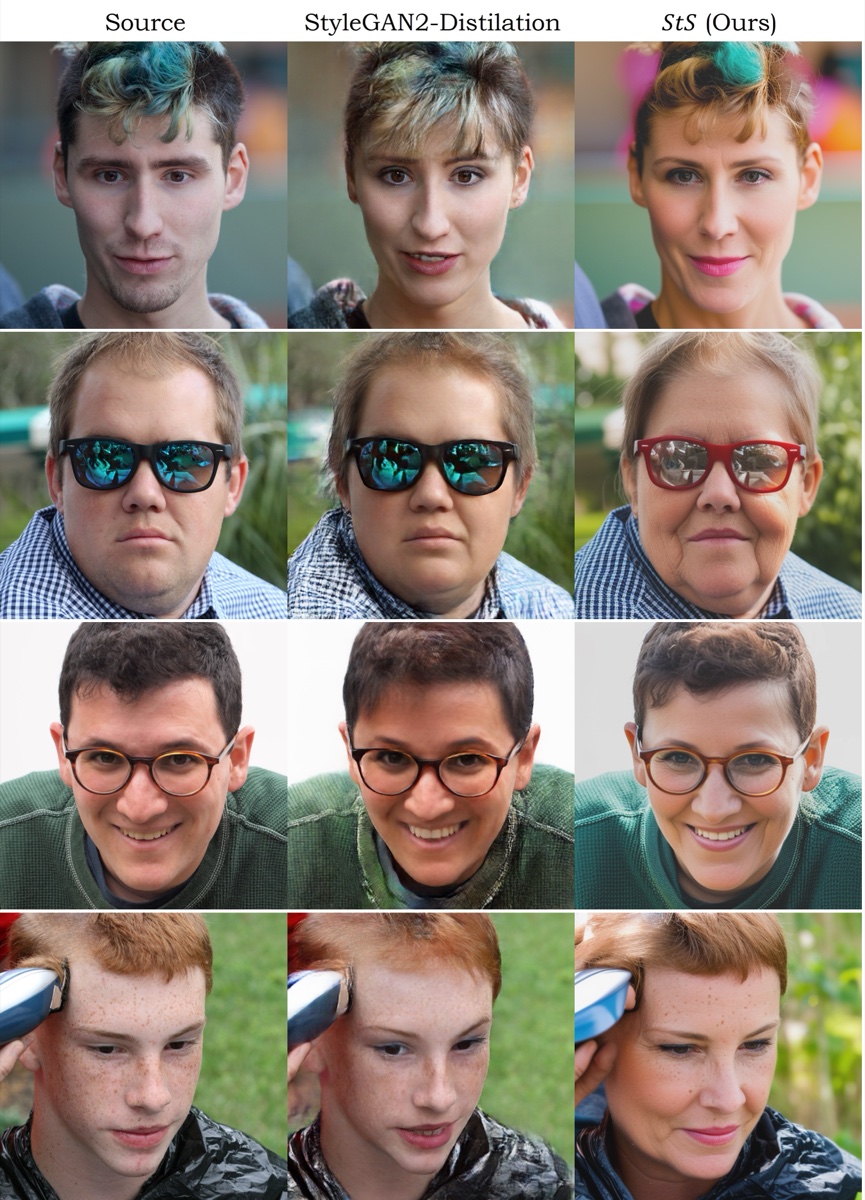}} \hfill
	\subfigure[To older]{\label{fig:subfigA2}\includegraphics[width=0.23\textwidth]{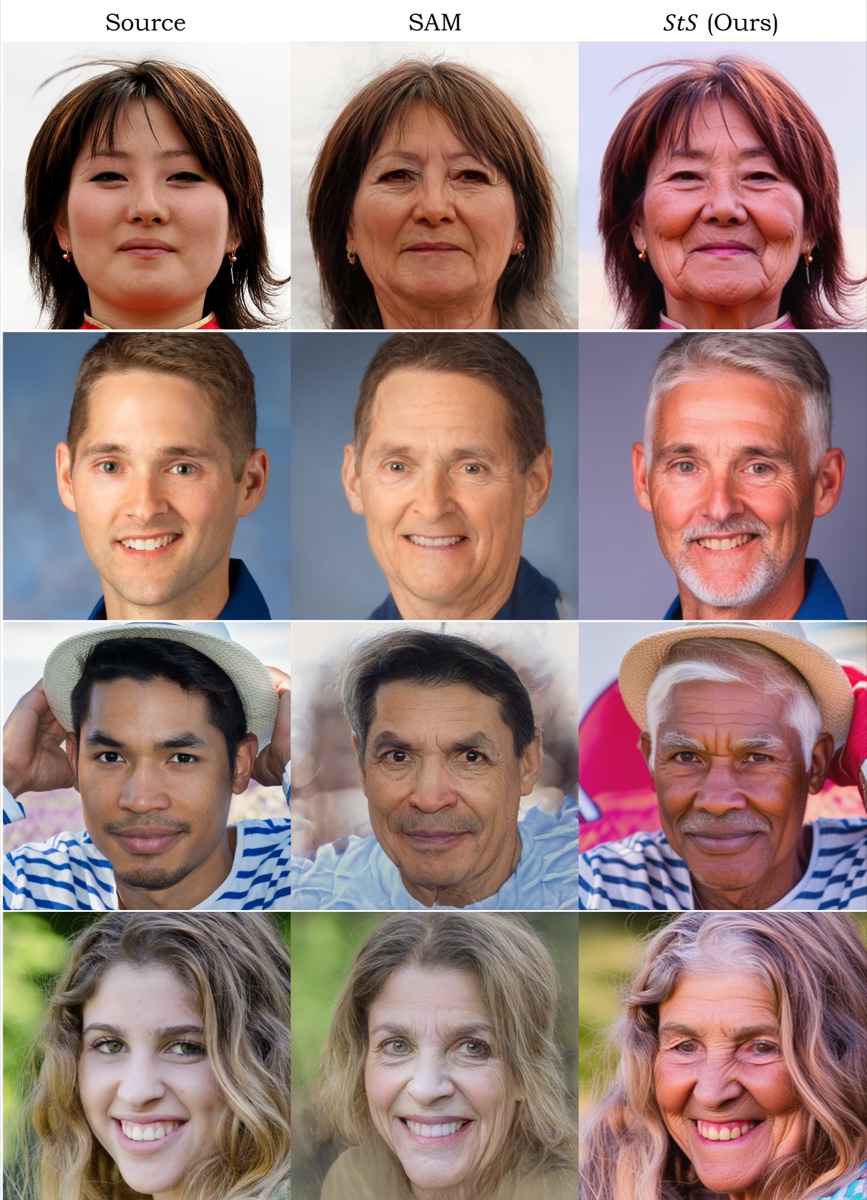}} \hfill
	\subfigure[To younger]{\label{fig:subfigB2}\includegraphics[width=0.23\textwidth]{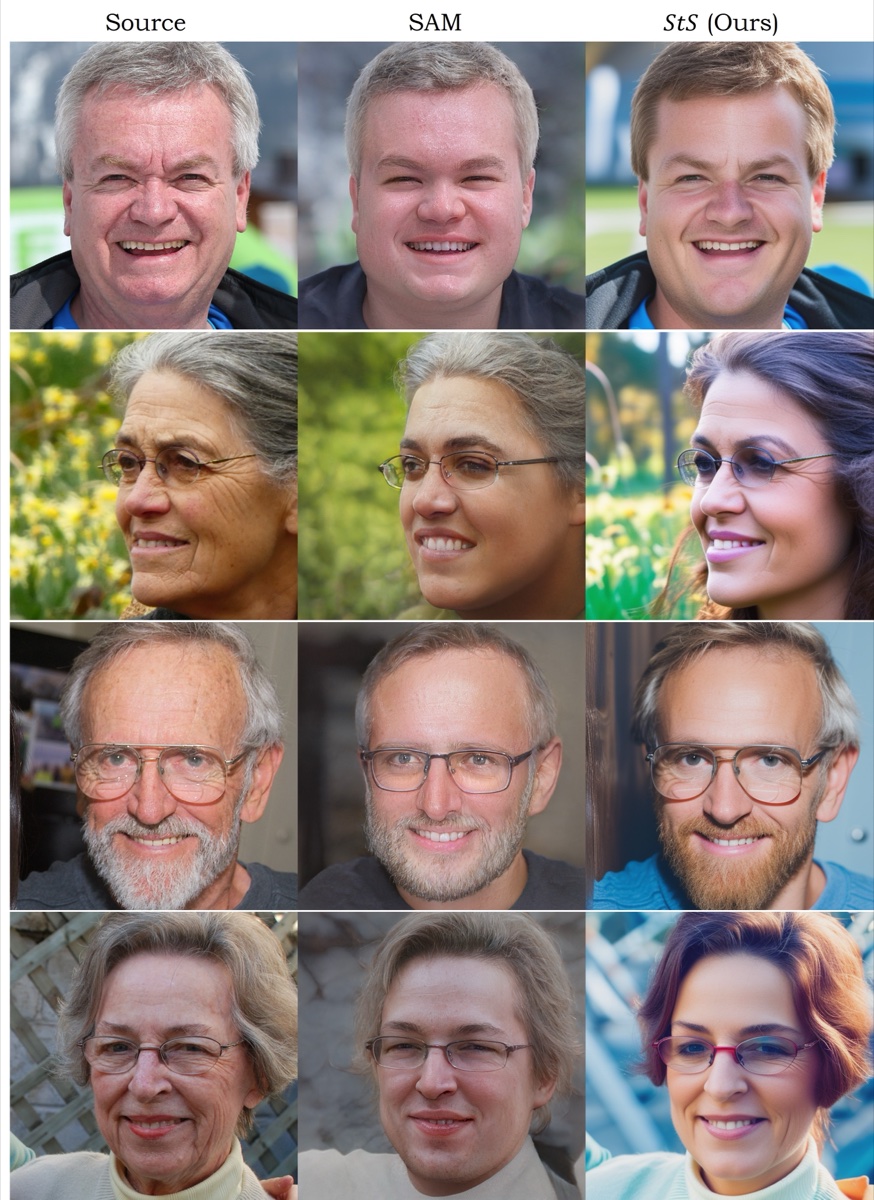}}
	\caption{\label{fig:combined}Additional applications: (a,b) Gender Swap, and (c,d) Age Translation over the \emph{FFHQ} dataset.}
\end{figure}

We emphasize that we claim no advantage for our model over existing diffusion-based editing techniques in cases where the source image is relatively simple (e.g., centered objects) or when the edits do not require close adherence to the source image. In such cases, existing diffusion-based methods produce high-quality results, and sometimes operate in a zero-shot manner (unlike our model, which requires task-specific optimization). Nevertheless, our model is versatile enough to achieve qualitatively competitive translations in these scenarios as well, compared to existing methods. \Cref{fig:combined} illustrates the performance of our \emph{StS} approach in gender-swap, compared to StyleGAN2-Distilation \cite{viazovetskyi2020stylegan2}, and in age translation, compared to SAM \cite{alaluf2021only}, respectively, both over the \emph{FFHQ} dataset.  In both gender-swap and age translation tasks, our model demonstrates competitive capabilities compared to the task-oriented baselines. As expected, our model adheres to the facial structure, pose, and expression of the source image, resulting in outputs that resemble a transformation of the input individual rather than depicting a different person from the target domain. This adherence is notably superior compared to the baselines in both tasks.

\subsection{I2IT with Flexible Adherence to Source Images}

It should be noted that since we use the Canny map as a spatial condition, our model is committed to preserving the object outlines present in the source image. Therefore, for example, \emph{StS} will not shorten the hair of a female input image when translating it into a male. This attribute limits our model to constrained translations, where the nature of the constraint is determined by the spatial condition provided to the ControlNet. This limitation is discussed in detail below.

While diffusion-based image editing and translation are common, as detailed in the literature review of the main paper, our model excels in cases that require strict adherence to the source image. Unlike the automotive and facial translation tasks discussed — where preserving the structure and semantics of the original image is crucial even in edited areas — some other image translation tasks require only minimal adherence in the modified regions. For instance, in common dog$\leftrightarrow$cat or horse$\leftrightarrow$zebra translations, the primary focus is on replacing one object with another while maintaining only the position of the original object.

\begin{figure}
    \centering
  \includegraphics[width=0.9\linewidth]{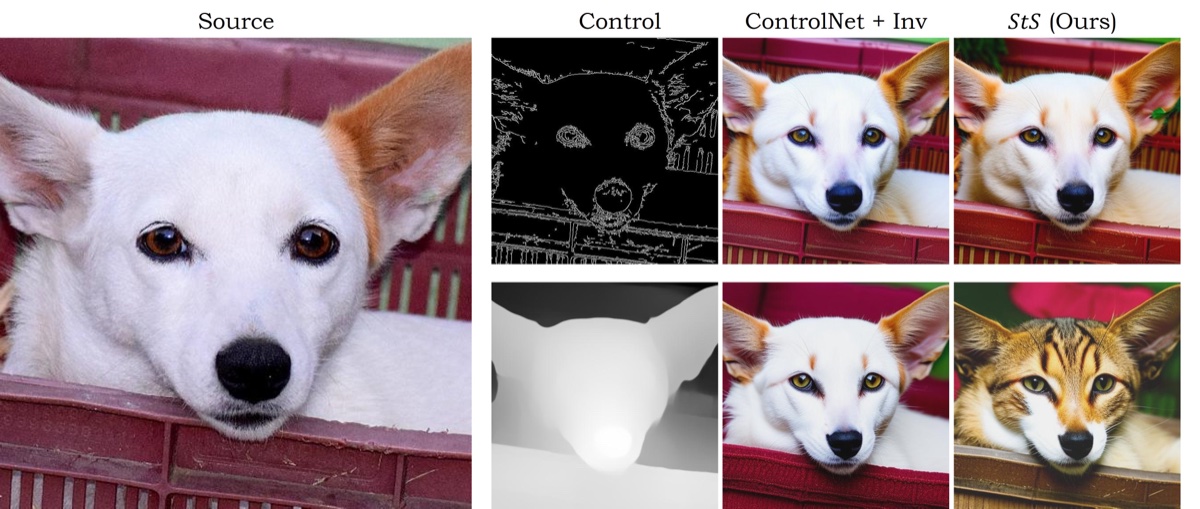}
    \caption{\textbf{Dog-to-Cat Translation.} A strict spatial constraint may hinder realism. Although a depth condition (bottom) is less restrictive than Canny (top), allowing the model to adjust the pattern of the fur to a more cat-like one, the boundaries still enforce a doglike structure (nose, eyes, ears, etc.)}
    \label{fig:doglike}
\end{figure}

Dogs and cats, for example, differ substantially from each other, so a dog translated from a cat is essentially just a dog posed in the same position as the source cat, without any further adherence to the cat's attributes. In such cases, the strict adherence of our model to the source image — expressed both in the seed space by the \emph{sts-GAN} and along the sampling trajectory by the ControlNet — may limit its ability to perform a realistic translation.

As discussed in Section 5 of the main text, the spatial constraint may restrict translation performance when the provided spatial control only crudely reflects the source domain. For example, in the dog$\leftrightarrow$cat translation task, \Cref{fig:doglike} illustrates a scenario where substantial structural modifications are necessary to transform a dog into a cat, creating a conflict with the spatial control.
In this example, the Canny control impeded the generation of the cat’s distinctive fur pattern, as ControlNet prioritized replicating the smooth fur edges of the source dog. Replacing the Canny control with a depth map (obtained using MiDaS \cite{Ranftl2022}) enabled the \emph{sts-GAN} to produce a more cat-like pattern. However, the distinct dog-like boundaries continued to conflict with the desired structural transformation. In such scenarios, the model may struggle to satisfy both spatial and appearance constraints, leading to unedited or unrealistic outputs.

In cases where the spatial control does not crudely reflect the source domain — such as when the boundaries of the source dog can be considered ``cat-like'' — our model performs the translation accurately, as demonstrated in \Cref{fig:dog_cat} using a depth map as the spatial control. It is important to note that while our model strives to closely adhere to attributes that can be preserved (e.g., expression), such strict adherence is not essential for these types of translations. Consequently, other image editing techniques may be more suitable for these kind of tasks.

\begin{figure}
    \centering
  \includegraphics[width=\linewidth]{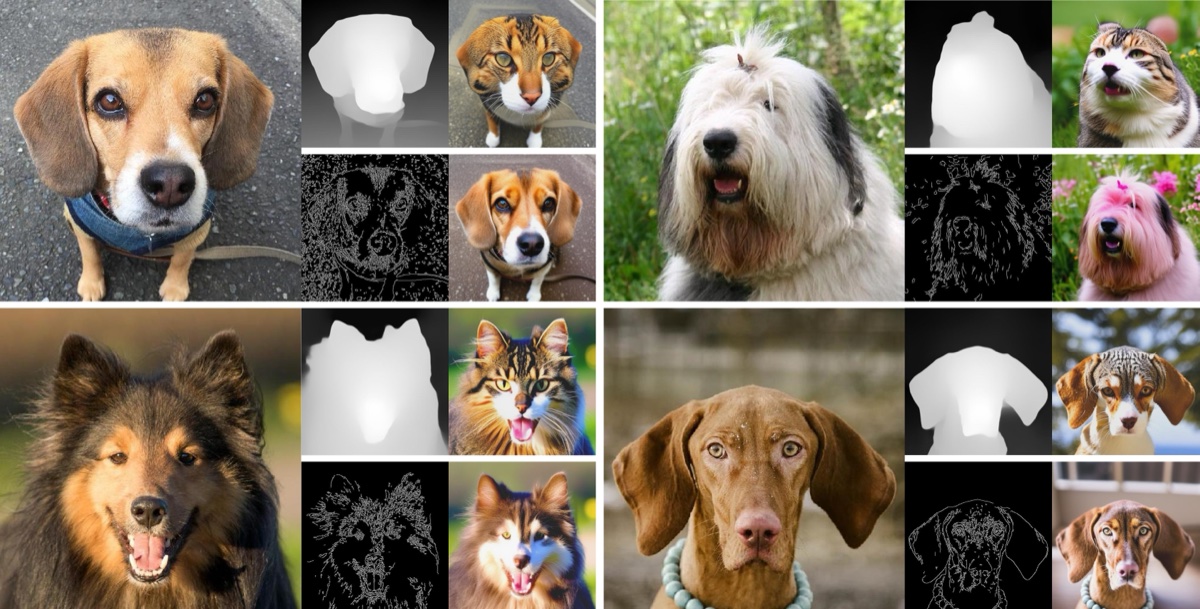}
    \caption{Example of failures in dog-to-cat translations. Each translation is shown for a ControlNet using depth (top) and Canny (bottom) conditions.}
    \label{fig:failure}
\end{figure}

\end{document}